\documentclass[journal]{IEEEtran}
\ifCLASSINFOpdf
\else
\fi
\usepackage{tikz}
\usetikzlibrary{shadows}
\usepackage{amsmath,amssymb}
\usepackage{bm}

\usepackage{multirow}
\usepackage{graphicx}
\usepackage{caption}
\usepackage{subcaption}
\usepackage{stfloats}
\usepackage{algorithm}
\usepackage{algorithmic}
\usepackage{adjustbox}

\begin{document}
%
% paper title
% Titles are generally capitalized except for words such as a, an, and, as,
% at, but, by, for, in, nor, of, on, or, the, to and up, which are usually
% not capitalized unless they are the first or last word of the title.
% Linebreaks \\ can be used within to get better formatting as desired.
% Do not put math or special symbols in the title.
\title{Modeling the Sequence of Brain Volumes by Local Mesh Models for Brain Decoding}

% author names and affiliations
% use a multiple column layout for up to three different
% affiliations
%\author{\IEEEauthorblockN{Itir Onal}
%	\IEEEauthorblockA{Department of Computer Engineering\\Middle East Technical University\\Ankara, Turkey\\
%		Email: itir@ceng.metu.edu.tr}
%	\and
%	\IEEEauthorblockN{Mete Ozay}
%	\IEEEauthorblockA{Graduate School of Information Sciences\\Tohoku University\\Sendai, Miyagi, Japan\\
%		Email: ozay.mete.b4@tohoku.ac.jp}
%	\and
%	\IEEEauthorblockN{Fatos T. Yarman Vural}
%	\IEEEauthorblockA{Department of Computer Engineering\\ Middle East Technical University\\ Ankara, Turkey
%		\\Email: vural@ceng.metu.edu.tr
%	}}

%\author{\IEEEauthorblockN{Itir Onal}
%\IEEEauthorblockA{Department Computer Engineering,
%Middle East Technical University,
%Ankara, Turkey\\ Email: itir@ceng.metu.edu.tr}
%}

%\author{\IEEEauthorblockN{Itir Onal\IEEEauthorrefmark{1},
%		Mete Ozay\IEEEauthorrefmark{2},
%		Ilke Oztekin\IEEEauthorrefmark{3},
%		Fatos T.Yarman Vural\IEEEauthorrefmark{1}}
%	\IEEEauthorblockA{\IEEEauthorrefmark{1}Department Computer Engineering,
%		Middle East Technical University,
%		Ankara, Turkey\\ Emails: \{itir,vural\}@ceng.metu.edu.tr}
%	\IEEEauthorblockA{\IEEEauthorrefmark{2}Graduate School of Information Sciences\\Tohoku University\\Sendai, Miyagi, Japan\\
%		Email: ozay.mete.b4@tohoku.ac.jp}
%	\IEEEauthorblockA{\IEEEauthorrefmark{3}Department of Psychology, Koc University, Istanbul, Turkey\\
%		Email: ioztekin@ku.edu.tr}
%}

\author{Itir~Onal,~\IEEEmembership{Member,~IEEE,}
	Mete~Ozay,~\IEEEmembership{Member,~IEEE,}
	Eda~Mizrak,
	Ilke~Oztekin,
	and~Fatos~T.~Yarman~Vural,~\IEEEmembership{Senior Member,~IEEE,}% <-this % stops a space
	\thanks{I. Onal and F. T. Yarman Vural are with the Department
		of Computer Engineering, Middle East Technical University, Ankara, Turkey.
		e-mails: \{itir,vural\}@ceng.metu.edu.tr}% <-this % stops a space
	\thanks{M. Ozay is with Graduate School of Information Sciences,Tohoku University,Sendai, Miyagi, Japan. e-mail: ozay.mete.b4@tohoku.ac.jp}% <-this % stops a space
	\thanks{E. Mizrak and I. Oztekin are with Department of Psychology, Koc University, Istanbul, Turkey. e-mails: \{emizrak, ioztekin\}@ku.edu.tr }}

% conference papers do not typically use \thanks and this command
% is locked out in conference mode. If really needed, such as for
% the acknowledgment of grants, issue a \IEEEoverridecommandlockouts
% after \documentclass

% for over three affiliations, or if they all won't fit within the width
% of the page, use this alternative format:
% 
%\author{\IEEEauthorblockN{Michael Shell\IEEEauthorrefmark{1},
%Homer Simpson\IEEEauthorrefmark{2},
%James Kirk\IEEEauthorrefmark{3}, 
%Montgomery Scott\IEEEauthorrefmark{3} and
%Eldon Tyrell\IEEEauthorrefmark{4}}
%\IEEEauthorblockA{\IEEEauthorrefmark{1}School of Electrical and Computer Engineering\\
%Georgia Institute of Technology,
%Atlanta, Georgia 30332--0250\\ Email: see http://www.michaelshell.org/contact.html}
%\IEEEauthorblockA{\IEEEauthorrefmark{2}Twentieth Century Fox, Springfield, USA\\
%Email: homer@thesimpsons.com}
%\IEEEauthorblockA{\IEEEauthorrefmark{3}Starfleet Academy, San Francisco, California 96678-2391\\
%Telephone: (800) 555--1212, Fax: (888) 555--1212}
%\IEEEauthorblockA{\IEEEauthorrefmark{4}Tyrell Inc., 123 Replicant Street, Los Angeles, California 90210--4321}}

% use for special paper notices
%\IEEEspecialpapernotice{(Invited Paper)}

% make the title area
\maketitle

% As a general rule, do not put math, special symbols or citations
% in the abstract
\begin{abstract}
	%Functional Magnetic Resonance Imaging (fMRI) measurements represent the Blood Oxygenation Level Dependent (BOLD) responses to a stimulus for about 10-12 seconds, in an event-related design experiment. This approach generates a brain volume for each time second. The smallest unit of this volume, called voxel, consists of the time series of intensity values which approximates the hemodynamic response (HDR). The huge amount of fMRI brain data is simplified by selecting the "most important" intensity values among the time series of "active" voxels. The massively interconnected and dynamic nature of human brain cannot be represented by considering only a collection of selected voxels and time instances, obtained from fMRI brain volumes. 
	%In experimental neuroscience, it is observed that spatially close voxels act together to generate similar responses to the same stimuli. Moreover, spatially remote voxels may exhibit functionally similar time series.  Following these observations, we assume a linear model among the time series recorded at each voxel of a the fMRI brain volumes in a locality defined by a neighborhood system. assuming a linear model among the BOLD responses.  
	
	We represent the sequence of fMRI (Functional Magnetic Resonance Imaging) brain volumes recorded during a cognitive stimulus by a graph which consists of a set of local meshes. The corresponding cognitive process, encoded in the brain, is then represented by these meshes each of which is estimated  assuming a linear relationship among the voxel time series in a predefined locality.
	
 First, we define the concept of locality in two neighborhood systems, namely, the \textit{spatial} and \textit{functional neighborhoods}. Then, we construct \textit{spatially} and \textit{functionally local meshes} around each voxel, called seed voxel, by connecting it either to its spatial or functional p-nearest neighbors. The mesh formed around a voxel is a directed sub-graph with a star topology, where the direction of the edges is taken towards the seed voxel at the center of the mesh. We represent the time  series recorded at each seed voxel in terms of linear combination of the time series of its p-nearest neighbors in the mesh. The relationships between a seed voxel and its neighbors are represented by the edge weights of each mesh, and are estimated by solving a linear regression equation. The estimated mesh edge weights lead to a better representation of information in the brain for encoding and decoding of the cognitive tasks. We test our model on a visual object recognition and emotional memory retrieval experiments using Support Vector Machines that are trained using the mesh edge weights as features. In the experimental analysis, we observe that the  edge weights of the spatial and functional meshes perform better than the state-of-the-art brain decoding models. 
	
\end{abstract}
\renewcommand\IEEEkeywordsname{Keywords}
% no keywords

% For peer review papers, you can put extra information on the cover
% page as needed:
% \ifCLASSOPTIONpeerreview
% \begin{center} \bfseries EDICS Category: 3-BBND \end{center}
% \fi
%
% For peerreview papers, this IEEEtran command inserts a page break and
% creates the second title. It will be ignored for other modes.
\IEEEpeerreviewmaketitle

\section{Introduction}
% no \IEEEPARstart

Functional Magnetic Resonance Imaging (fMRI) is used as the primary modality to capture neural activations in the brain due to its high spatial resolution and reasonable temporal resolution \cite{Kefayati2013}. It measures the Blood Oxygenation Level Dependent (BOLD) responses to a stimulus for about 10-12 seconds in an event-related design experiment. This approach generates a brain volume for approximately each time instance. The smallest unit of this volume, called voxel, consists of the time series of intensity values which approximate the hemodynamic response (HDR). 

The techniques employed to learn the brain activity patterns from the BOLD signals are called Multi Voxel Pattern Analysis (MVPA). Following the pioneering study of Haxby et al. \cite{Haxby2001}, machine learning techniques have been used for  MVPA  for diagnosing disorders \cite{Kloppel2012,Orru2012}, hypothesis validation \cite{Oztekin2011} and classifying cognitive states which is a method for predicting cognitive states called brain decoding. In the context of machine learning, encoding the brain activities is formalized by training a classifier using a training set of fMRI measurements, whereas a cognitive state is decoded by assigning a label to a test fMRI measurement recorded during one of the prescribed cognitive stimuli.

In state-of-the-art MVPA approaches, cognitive states are usually represented by concatenating the selected voxel intensity values to construct a vector in a feature space. The time series recorded for each voxel can be represented by various methods, such as computing maximal or mean values. Also, some of the inactive voxels can be eliminated by Principal Component Analysis, Independent Component Analysis, Searchlight and GLM analysis to reduce the dimension of the feature space. Then, a classifier such as Support Vector Machine (SVM), k-Nearest Neighbor (k-NN) or Naive Bayes is employed using the feature vectors. 

While some of the state-of-the-art methods \cite{Cox2003,Kamitani2005,Kamitani2006,Ng2010,Ng2011} are designed to extract only spatial patterns from the BOLD responses, the others \cite{Mitchell2004,Ng2011_2} focus on modeling temporal information to form features for brain decoding. A popular method for extracting a spatial feature for brain decoding from fMRI data is computing the average BOLD response for each voxel to represent a stimulus \cite{Cox2003,Kamitani2005,Kamitani2006}. A recent work, suggested by  \cite{Ng2010,Ng2011}, considers each brain volume  as a sample. On the other hand, Mitchell et al. \cite{Mitchell2004} and Ng et al. \cite{Ng2011_2} concatenate the BOLD responses within the same trial to extract features that include spatio-temporal information. 
%Although they represent spatial or spatio-temporal information  embedded in the voxels, none of the aforementioned studies propose a model to formalize the relationships among the brain volumes, recorded along the time course of a cognitive stimulus. In other words, spatial models mostly lack the temporal information while the temporal models loose most of the spatial relationships among the voxel BOLD responses.  {\color{red} (BU CUMLE ACIK DEGIL ONLAR DA ILISKILERI FORMALIZE EDIYOR, BIZIM FARKIMIZ DAHA ACIK ANLATILMALI)}.

In order to extract the temporal information from the fMRI data, various studies \cite{Pantazatos2012,Richiardi2011,Firat2014,Baldassano2012,Firat2013_b} compute the pairwise correlations between the responses of voxels or brain regions. Among these studies, Pantazatos et al. \cite{Pantazatos2012} model temporal relationships by using pairwise correlations between brain regions as features for brain decoding. Moreover, Richiardi et al. \cite{Richiardi2011} model a functional connectivity graph using the pairwise correlations between brain regions, and they decode the brain states using graph matching algorithms. Also, edge weights of their proposed brain graphs are generally selected using the correlation between pairs of selected voxels, and vertices or edge weights of these brain graphs are used as features for cognitive state classification \cite{Richiardi2013}. Contrary to the coarse-level functional connectivity, Firat et al. \cite{Firat2014} use pairwise correlations of voxels as features while Baldassano et al. \cite{Baldassano2012} extract the functional connectivity structures by modeling pairwise connectivity among voxels.  Also, a minimum spanning tree of a graph structure defined using a functional connectivity matrix is employed using all voxels for each stimulus for brain decoding \cite{Firat2013_b}. Note that, the aforementioned temporal methods are designed to model only the pairwise relationships between voxels and employ these  relationships for brain decoding.  Although these methods represent the spatio-temporal information to a certain degree, none of them propose a model to formalize the relationships among the brain volumes, recorded along the time course of a cognitive stimulus. In other words, spatial models mostly lack the temporal information while the temporal ones lose  the spatial relationships among the voxel BOLD responses.  %{\color{red} (BU CUMLE ACIK DEGIL ONLAR DA ILISKILERI FORMALIZE EDIYOR, BIZIM FARKIMIZ DAHA ACIK ANLATILMALI)}.

The major motivation of this study follows the observations of the functional and spatial similarities of voxel BOLD responses in a pre-defined locality. As an example, Bazargani et al. \cite{Bazargani2014} state that neighboring voxels belonging to a homogeneous ROI have Hemodynamic Response Function (HRF) with the same shape, possibly with slightly varying amplitude. In other words, spatially close voxels tend to give similar BOLD responses to the same stimuli.  Moreover, Kriegeskorte et al. \cite{Kriegeskorte2006} report that a univariate model of activations fails to benefit from local spatial combination of signals, and performs worse than Searchlight methods for detection of informative regions. They observe that spatially remote voxels may also exhibit functionally similar time series. 

Furthermore, Ozay et al. \cite{Ozay2011} observe that the voxel time series do not differ significantly to discriminate the different cognitive states, but there exist slight variations among the voxel intensity values of voxels located in a spatial neighborhood. They propose a set of Local Meshes (LMM) to model the relationship among the spatially close voxels. They show that the  LMM features perform better than the voxel intensity values for the classification of cognitive states. However, in \cite{Ozay2011}, only the brain volume which is obtained 6 seconds after the stimulus presentation is used for the construction of a model while the remaining ones are discarded under the canonical HRF assumption.

Finally, Firat et al. \cite{Firat2013} propose a Functional Mesh Model (FMM) which is used to construct a set of local meshes by selecting the nearest neighbors of voxels defined within a functional neighborhood. Their experimental results show that their features are more discriminative than the features obtained within a spatial neighborhood. Yet, they also estimate the mesh weights using only a single intensity value for each voxel. 

In none of the above mentioned models, the time series recorded at all the voxels are fully employed to represent the spatial relationships among the voxels of the fMRI data. The available approaches either use the active voxels, the most crucial time instances or both. Although these approaches smooth the noise in voxel time series, and represent the huge amount of data in a more compact way, it may result in losing important information embedded in time series of brain volumes, recorded under a stimulus. 

The massively interconnected and dynamic nature of human brain cannot be represented by considering only a collection of selected voxels and/or time instances which are obtained from fMRI data.  In addition, the above mentioned studies indicate that the relationship among the voxels is more informative than the information provided by the individual voxels. Therefore, it is desirable to develop a model which represents the relationship among the brain volumes, recorded during a stimulus. Furthermore, if this model incorporates the local similarities among the voxel time series, then it is possible to represent the sequence of brain volumes, recorded during a cognitive stimulus, by a graph which consists of a set of subgraphs each of which is represented  by a local linear regression model.

In this study, we employ the voxel time series measured during a cognitive task to create a local mesh around each voxel which models the relationship among the BOLD responses. The concept of locality is defined over a neighborhood system.  In this study, we introduce two types of neighborhoods, namely, spatial and functional neighborhoods.  The local mesh around each voxel is constructed by using either a spatial \cite{Onal2015_1} or functional neighborhood system to generate a set of spatially or functionally local meshes. We form a local mesh around each voxel by connecting the voxel to its p-nearest neighbors under a star topology. In each mesh, BOLD response of a voxel is represented by a weighted linear combination of BOLD responses of its spatially or functionally nearest neighbors. The weights are estimated by solving a linear regression equation, and are assigned to the edges of the mesh. The mesh edge weights estimated for each voxel enable us to represent the fMRI brain volumes recorded during a stimulus by a directed graph which consists of an ensemble of local meshes. 

The proposed local mesh models are different from the connectivity models defined over pairwise similarities between voxels and/or regions in the sense that the connectivity is defined over a local neighborhood of voxels. We employ the estimated weights of mesh edges as features to train and test the state-of-the-art Support Vector Machines (SVM) for encoding and decoding a set of cognitive processes measured by fMRI signals. 
 
Our approach is employed for modeling the sequence of fMRI brain volumes, recorded during an event-related design experiment. We assume that the voxel time series approximate the hemodynamic response for each stimulus. Unfortunately, block design lacks this information due to the overlaps between consecutive hemodynamic responses. Therefore, we test our approach in two event-related design experiments namely visual object recognition and emotional memory retrieval experiments. We observe that the classifiers which employ the proposed  local mesh ensembles outperform the state of the art MVPA models.  

\begin{figure}[t!]
	\centering
	%\shorthandoff{=}
	\includegraphics[scale=0.36]{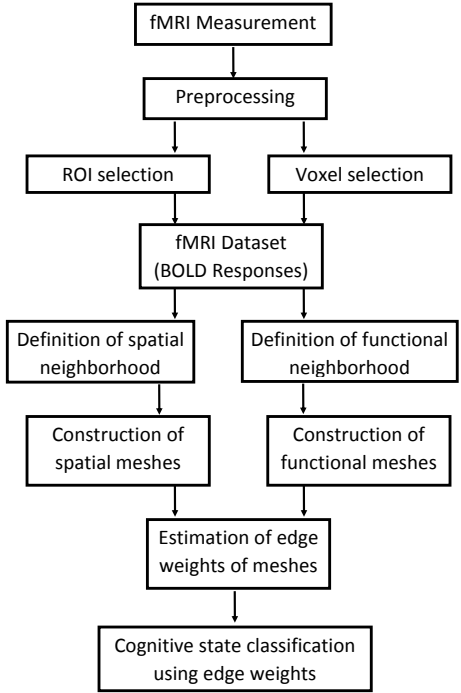}
	%\shorthandon{=}
	\caption{
	The flowchart of the proposed approach for obtaining local meshes and classification of cognitive states. }
	\label{fig:flowchart}
\end{figure}

In Fig.~\ref{fig:flowchart}, the steps of the proposed method are summarized. First, we preprocess the fMRI measurements and obtain voxel intensity values. Then, we select voxels or ROI to eliminate the redundant voxels and to reduce the noise in our dataset. After that, we define spatial and functional neighborhood systems, and construct meshes, accordingly. Finally, we estimate edge weights of meshes,  and employ them as features for cognitive state classification.

\section{Neuroimaging Data Collection}

\subsection{Visual Recognition Experiment}

%\begin{figure}
%	\centering
%	%\shorthandoff{=}
%	\includegraphics[scale=0.3]{experiment_object.png}
%	%\shorthandon{=}
%	\caption{A sample period from visual object recognition fMRI experiment. In each trial participant is presented with a gray scale image belonging to either bird or flower category for 4 seconds. Then, a rest period whose duration varies randomly among 8, 10 or 12 seconds follows this presentation to ensure that different samples are not correlated due to hemodynamic response function. The participant is expected to recognize whether the current object belongs to the same category as the previous one. If the current and previous objects belong to same category, then the user is expected to press the button (indicated by red arrow).}
%	\label{fig:experiment_object}
%\end{figure}

In this experiment, fMRI measurements wee recorded while participants performed a one-back repetition detection task. The stimuli consisted of gray-scale images belonging to two categories, namely, birds and flowers. Each stimulus was presented for 4 seconds, and followed by a rest period of 8 seconds.

%	\begin{figure}
%		\centering
%		%\shorthandoff{=}
%		\includegraphics[scale=0.34]{experiment_object.png}
%		%\shorthandon{=}
%		\caption{A sample period from visual object recognition fMRI experiment. In each trial participant is presented with a gray scale image belonging to either bird or flower category for 4 seconds. Then, a rest period whose duration varies randomly  among 8, 10 or 12 seconds follows this presentation. The participant is expected to recognize whether the current object belongs to the same category as the previous one. If the current and previous objects belong to same category, then the user is expected to press the button (indicated by red arrow).}
%		\label{fig:experiment_object}
%	\end{figure}

\subsection{Emotional Memory Retrieval Experiment}

In this experiment, the stimuli consisted of two neutral (Kitchen utensils and Furniture) and two emotional (Fear and Disgust) categories of images. Each trial started with a 12 seconds fixation period followed by a 6 seconds encoding period. In the encoding period, participants were presented with 5 images from the same category, each image lasting 1200 on the screen. Following the fifth image, a 12 seconds delay period was presented in which participants solved three math problems consisting of addition or subtraction of two randomly selected two-digit numbers. Following the third math problem, a 2 seconds retrieval period started in which participants were presented with a test image from the same category and indicated whether the image was a member of the current study list or not. For a similar experimental setting,  please refer to \cite{Mizrak2015}. For classification, we employed measurements obtained during the encoding and retrieval phase as our training and test data, respectively.

\section{Pre-processing of Analysis of Neuroimaging Data}

SPM8 was used for pre-processing and data analysis\footnote{http://www.fil.ion.ucl.ac.uk/spm/}. Preprocessing of images consisted of (a) correction of slice acquisition timing across slices, (b) realigning the images to the first volume in each fMRI run to correct for head movement, (c) normalization of functional and anatomical images to a standard template EPI provided by SPM2, and (d) smoothing images with a 6-mm full-width half-maximum isotropic Gaussian kernel.
Finally, we extract 116 Automated Anatomical Labeling (AAL) regions \cite{Tzourio2002} using Marsbar \cite{Brett2002}. For the visual object recognition experiment, active voxels were identified using the General Linear Model implemented in SPM and occipital lobe was selected as our region of interest (ROI). The task engaged visual cortex activity, where pattern analyses were employed. For the emotional memory retrieval experiment we selected first 3000 most discriminative voxels from the whole brain analysis, using ReliefF \cite{RobnikSikonja2003}.

In the visual object recognition experiment, 36 measurements are recorded in each of the 6 runs, and there are 216 samples. We design our datasets by taking the first 12 samples per each class from each run for training, and the last 6 samples per each class from each run for validation and testing. On the other hand, for the emotional memory retrieval experiment, 35 measurements are recorded within each run. Measurements recorded during the encoding phase are employed as training data and the ones obtained during retrieval phase are used as validation and test data.

\begin{figure*}[t]
	\centering
	\begin{subfigure}[b]{0.41\textwidth}
		\includegraphics[width=\textwidth]{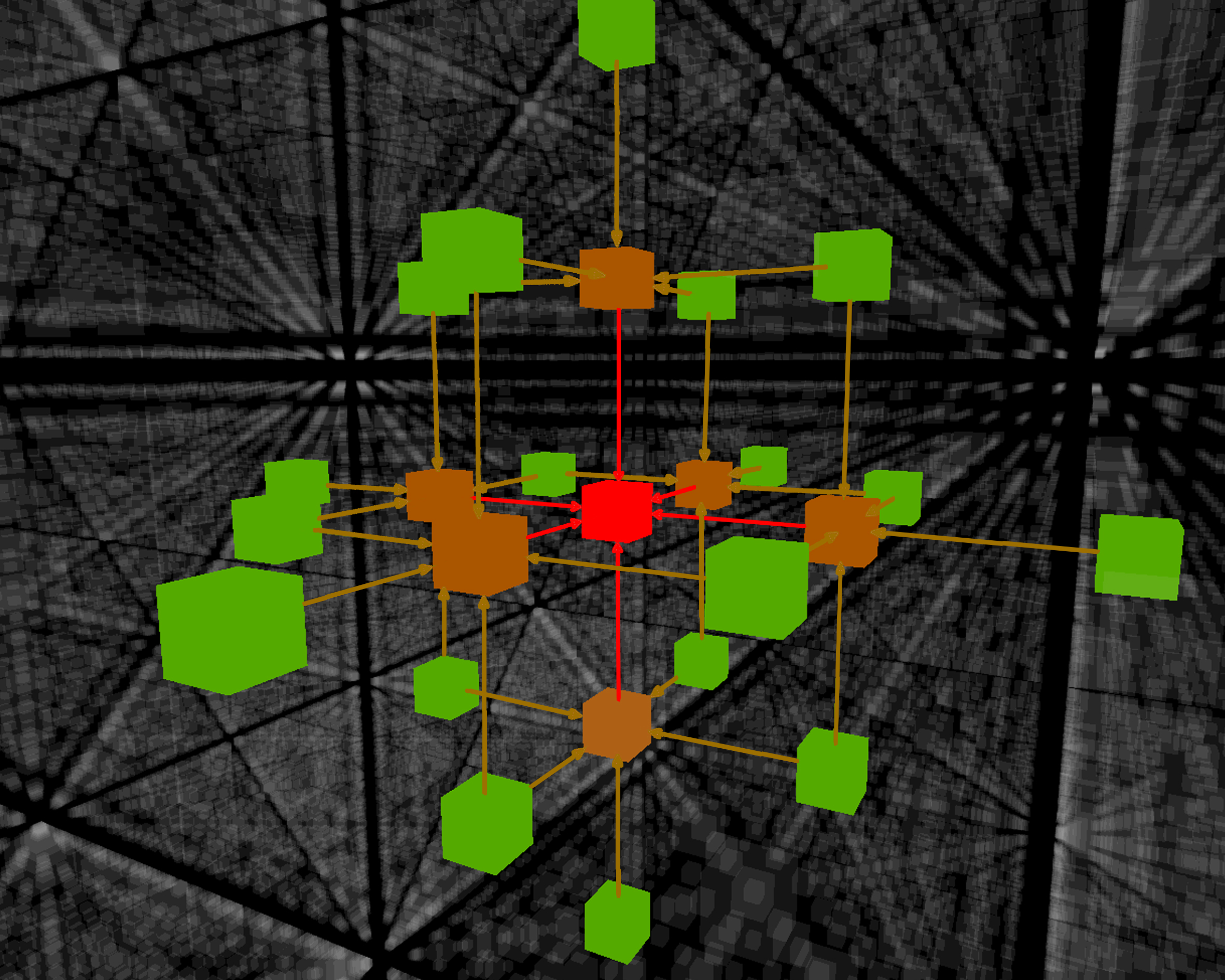}
		\caption{An ensemble of spatially local meshes. }
	
		\label{fig:ensemble_spatial}
	\end{subfigure}
	~ %add desired spacing between images, e. g. ~, \quad, \qquad, \hfill etc. 
	%(or a blank line to force the subfigure onto a new line)
	\begin{subfigure}[b]{0.41\textwidth}
		\includegraphics[width=\textwidth]{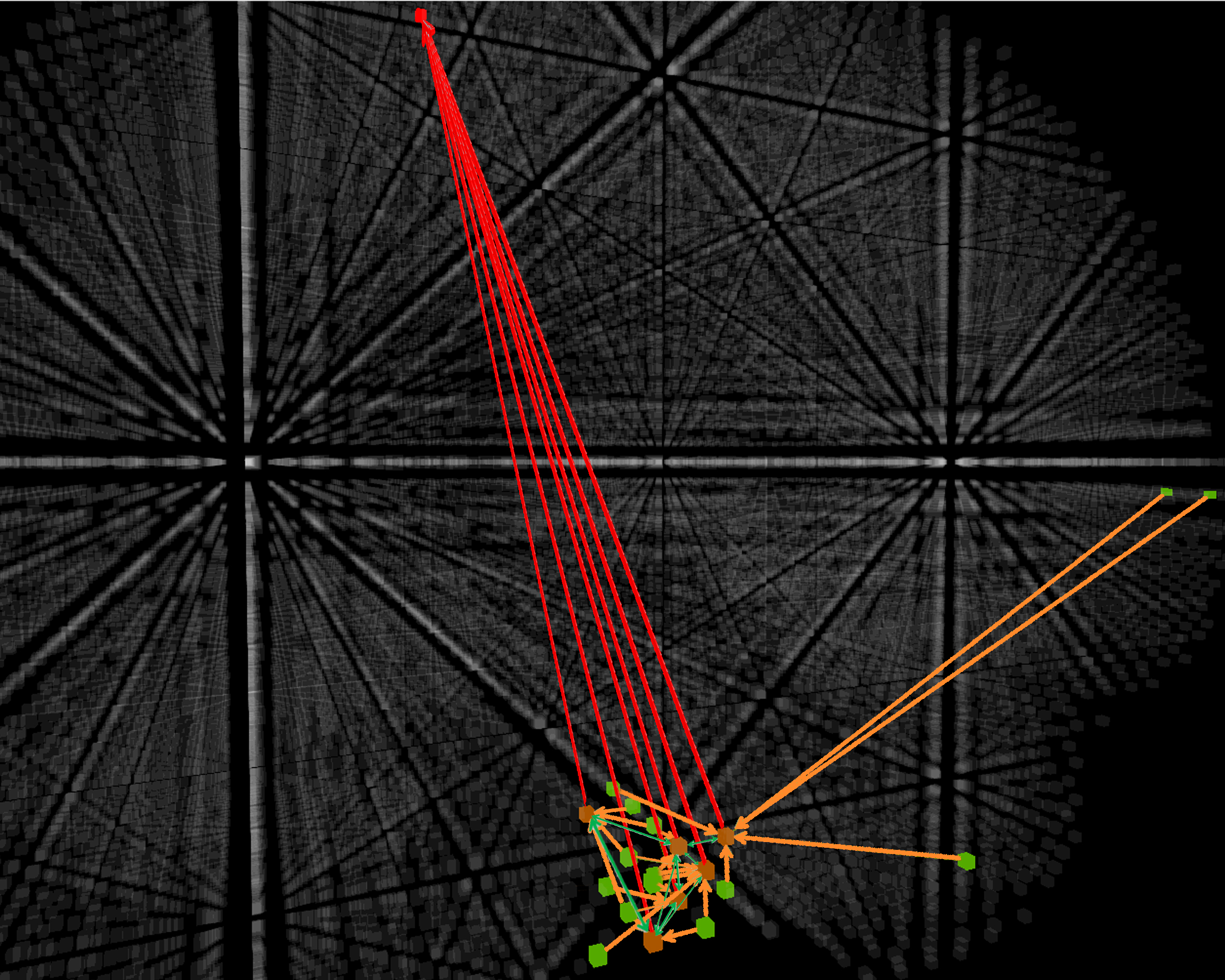}
		\caption{An ensemble of functionally local meshes.}
		\label{fig:ensemble_functional}
	\end{subfigure}
	~ %add desired spacing between images, e. g. ~, \quad, \qquad, \hfill etc. 
	%(or a blank line to force the subfigure onto a new line)

	\caption{Ensemble of meshes formed using (a) spatial neighbors and (b) functional neighbors of seed voxels. In both of the ensembles, the voxels denoted with red color are the seed voxels. Orange voxels denote the spatially nearest neighbors of the red voxel in (a), and the functionally nearest neighbors of the red voxel in (b). Green voxels are the spatially nearest neighbors of the corresponding orange voxels in (a), and functionally nearest neighbors of the corresponding orange voxels in (b). Only the highest edge weights are displayed for simplicity. }\label{fig:ensemble_of_meshes}
\end{figure*}

\section{Spatial and Functional Local Meshes}

The local mesh model proposed in this study is constructed using a collection of voxels located within a neighborhood of each voxel. For this purpose, first we make a formal definition of locality. This task is achieved by defining spatial and functional neighborhood systems. Then, we define two types of local meshes, namely, spatial and functional meshes. 

In the following sub-sections, we provide the formal definition of neighborhood systems and local meshes.   

\subsection{Neighborhood Systems}

One of the most important tasks of this study is to construct a linear relationship among the BOLD responses of voxels by employing ``locality" properties of the voxels. In order to achieve this goal, we define two types of neighborhood systems, namely, spatial and functional neighborhoods.

\textbf{Definition 1: Spatial Neighborhood:} Let $V =\{ v_j\}$  be a set of nodes each of which  corresponds to a voxel located at a coordinate $\bar{l_j}$. Spatially nearest neighbor of a voxel $v_j$  is defined as  the voxel $v_k$ which has the smallest Euclidean distance to the seed voxel $v_j$, as given below:
\begin{equation}
\label{eq:n1}
\eta_1^{spat}[{v_j}] = \{v_k: \|\bar{l_j} - \bar{l_k} \| \leq \|\bar{l_j} - \bar{l_o} \|,   \forall  \bar{l_o} \}.
\end{equation}

$p$-spatially nearest neighbors of voxel $v_j$ located at $\bar{l}_j$ are defined recursively as
\begin{multline}
\label{eq:np}
\eta_p^{spat}[v_j] =  \{v_k \cup \eta_{p-1}^{spat}[v_j] :  \|\bar{l_j} - \bar{l_k} \| \leq \|\bar{l_j} - \bar{l_o} \|, \\  \forall v_o \in \eta^{spat}_{p-1}[v_j]^c  \}  ,
\end{multline}
where $\eta^{spat}_{p-1}[v_j]^c$ is the set complement of the set $\eta^{spat}_{p-1}[v_j]$.
 
In the above definition of $p$-spatially nearest neighborhood, when $p = 6$,  6-nearest neighbors of a voxel correspond to the adjacent voxels at the right, left, up, down, front and back of that voxel (see Fig. \ref{fig:ensemble_spatial}). As we increase \textit{p}, the set of the neighboring voxels gets larger. The voxels  which are adjacent to the nearest neighbors are included recursively, with an increasing $p$.

\begin{figure*}[ht]
	\centering
	%\shorthandoff{=}
	\includegraphics[scale=0.48]{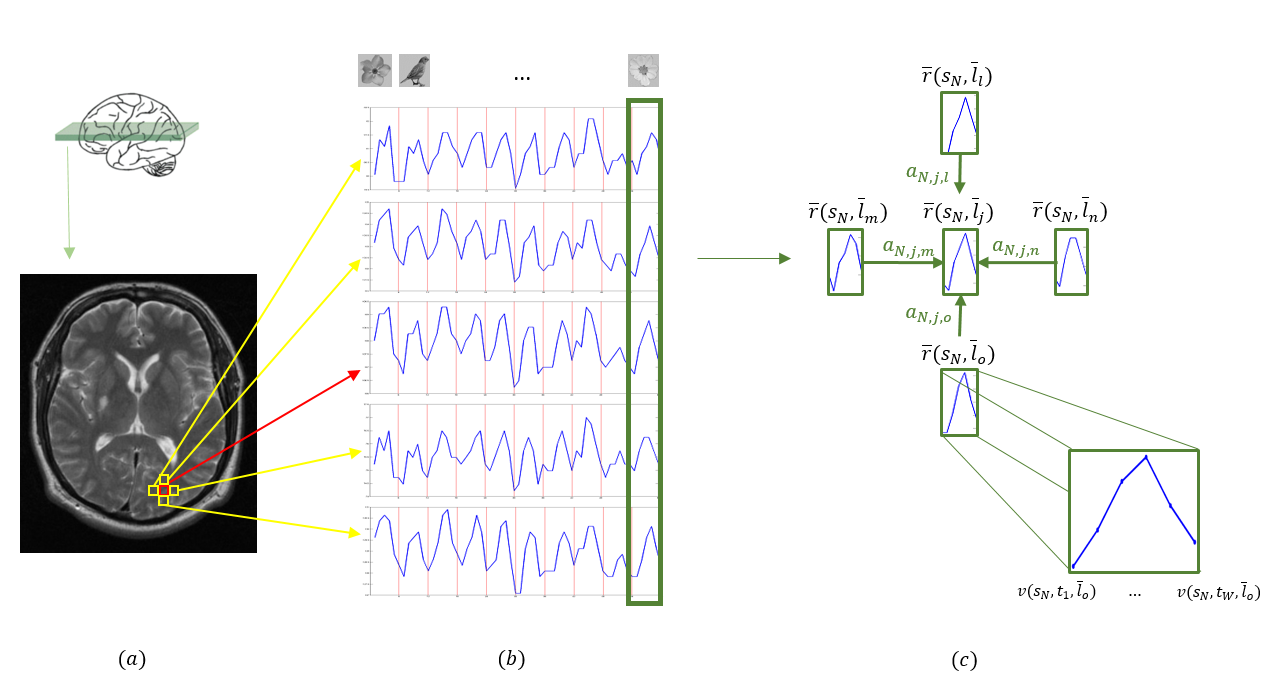}
	%\shorthandon{=}
	\caption{
		Design of a spatially local mesh for modeling the relationship among the BOLD responses of voxels. (a) For simplicity, a seed voxel is depicted in an axial slice at the center of a mesh (red) with its four spatially nearest neighbors (yellow). Neighboring voxels are selected within a 3D brain volume. For example, if $p=6$, then we would include the voxels above and below the seed voxel. (b) fMRI time series are recorded at each voxel, for each stimulus for bird and flower classes. (c) A local mesh is computed using the responses obtained from a seed voxel $\bar{r}(s_N, \bar{l_j})$ and its four nearest neighbors for the last sample $s_N$. We also visualize voxel intensity values of a sample response $\bar{r}(s_N, \bar{l_o})$ obtained from a voxel  at $\bar{l_o}$.}
	\label{fig:lmm-tm}
\end{figure*}

\begin{figure*}[t]
	\centering
	%\shorthandoff{=}
	\includegraphics[scale=0.48]{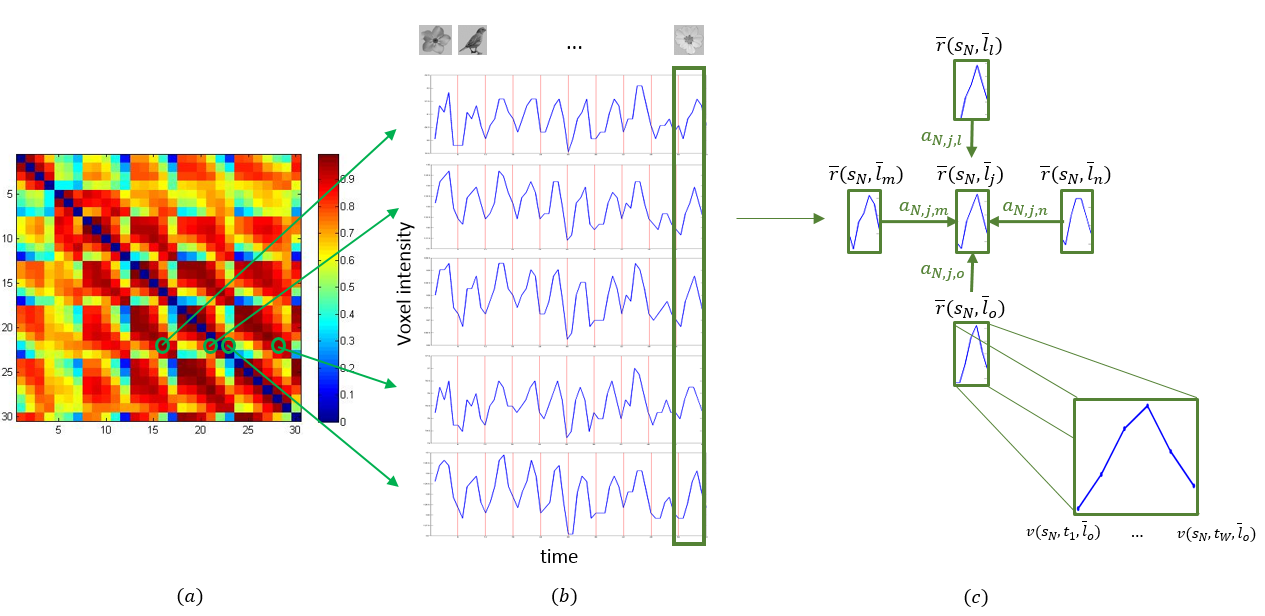}
	%\shorthandon{=}
	\caption{ Design of a functionally local mesh using the voxel time series. (a) A sample functional connectivity matrix formed using Pearson correlation. p-functionally nearest neighbors of a voxel are denoted as the ones with the darkest red color. (b) fMRI measurements are recorded for each stimulus for bird and flower classes. (c) We depict a functional mesh formed around a seed voxel, in star topology. The direction of the edges are taken towards the seed voxel, indicating that the seed voxel is represented by the linear combination of its neighbors. The edge weights are computed by using the response obtained from a seed voxel $\bar{r}(s_N, \bar{l_j})$ and the responses of its four functionally nearest neighbors,  $\bar{r}(s_N, \bar{l_i})$ ,  $\bar{r}(s_N, \bar{l_n})$,  $\bar{r}(s_N, \bar{l_m})$,  $\bar{r}(s_N, \bar{l_o})$, for the last sample $s_N$. We also visualize voxel intensity components of a sample response $\bar{r}(s_N, \bar{l_o})$ obtained from a voxel located at $\bar{l_o}$.}
	\label{fig:fmm-tm}
\end{figure*}

\textbf{Definition 2: Functional Neighborhood:} Let $R(v_j)$ denote the vector with the entries of BOLD signal measurements at  voxel $v_j$ which are recorded as the response of a series of stimulus during the entire experiment. A functionally nearest neighbor of a voxel $v_j$, is defined as the voxel $v_k$ when the vector $R(v_k)$ is maximally correlated with  $R(v_j)$ measured at voxel $v_j$. In order to find the functionally nearest neighbor of a voxel $v_j$,  Pearson correlation between $R(v_j)$ and $R(v_k)$ is computed  for all the voxels $v_k$ in the brain volume using the following criterion:
\begin{equation}
\label{eq:correlation}
cor(R(v_j),R(v_k))= \frac{cov(R(v_j),R(v_k))}{\sigma(R(v_j)) \sigma(R(v_k))},
\end{equation}
where covariance between responses $R(v_j)$ and $R(v_k)$ is,

\begin{multline}
\label{eq:covariance}
cov(R(v_j),R(v_k))= \frac{1}{N*D} * \\ \sum_{\forall i,d}\left(v(s_i, t_d, \bar{l_j}) - \mu(R(v_j)) \right) * \left(v(s_i, t_d, \bar{l_k}) - \mu(R(v_k)) \right).
\end{multline}

and standard deviation of response $\sigma(R(v_j))$ is defined as 

\begin{equation}
\label{eq:std}
\sigma(R(v_j))= \sqrt{\sum_{\forall i,d}\left(v(s_i, t_d, \bar{l_j}) - \mu(R(v_j)\right)^2};
\end{equation}

and mean of response $\mu(R(v_j)$ is defined as

\begin{equation}
\label{eq:mean}
\mu(R(v_j))= \frac{1}{N*D} \sum_{\forall i,d}v(s_i, t_d, \bar{l_j});
\end{equation}

Then, a functionally nearest neighbor of a voxel $v_j$ is defined as:
\begin{multline}
\label{eq:n1}
	\eta_1^{func}[{v_j}] = \{v_k: cor(R(v_j),R(v_k)) \leq cor(R(v_j),R(v_o)), \\  \forall  v_o \}.
\end{multline}

Finally, p-functionally nearest neighbors of the voxel $v_j$ is recursively defined as follows;
\begin{multline}
\label{eq:np}
\eta_p^{func}[v_j] =  \{v_k \cup \eta_{p-1}^{func}[v_j] : \\ cor(R(v_j),R(v_k)) \leq cor(R(v_j),R(v_o)),   \forall v_o \in \eta^{func}_{p-1}[v_j]^c  \} .
\end{multline}

To this end, we select the functional neighbors of the voxel $v_j$ as the voxels having \textit{p} of the highest Pearson correlation with that voxel.

Note that functionally $p$-nearest neighbors of a voxel may or may not be the same as the spatially $p$-nearest neighbors. Practical evidence indicates that most of the spatially close voxels are also functionally close, i.e., they are highly correlated. However, some of the voxel pairs are spatially far apart, yet they are functionally close to each other (see Fig. \ref{fig:ensemble_functional}).

% As another similarity measure, we employed cosine similarity between the responses of each voxel pair using
%\begin{align}
%\label{eq:cosine}
%cos(X,Y)= \frac{X  Y^T}{ \left( X  X^T \right) \left( Y  Y^T \right) }.
%\end{align}	
%In order to form a functional neighborhood using cosine similarity $(\eta^{cos}_p)$, we select $p$ voxels whose cosine similarity values computed with respect to the seed voxel are the maximum among others. Finally, we computed Euclidean distance between all pairs of voxel BOLD responses during training and formed a functional neighborhood $(\eta^{euc}_p)$. We computed Euclidean distance between two BOLD responses using
%\begin{align}
%\label{eq:euclidean}
%euc(X,Y)= \sqrt{ (X - Y)  (X - Y)^T}.
%\end{align} 
%$p$ voxels, whose BOLD responses over all training data have the smallest Euclidean distance with that of the seed voxel, are selected as the functionally nearest neighbors. To this end, we select functional neighbors as the ones having the highest Pearson correlation, the highest cosine similarity and the lowest Euclidean distance with the seed voxel.

\subsection{Construction of Local Meshes}

Based upon the neighborhood systems introduced in the previous section, the concept of locality is represented by a set of meshes defined over a sequence of brain volumes. For this purpose, two types of meshes, namely, spatially and functionally  local meshes, are established.

Both types of meshes are constructed around each voxel of the brain volume using a predefined neighborhood system. Once we select the neighborhood system, we form a local mesh around each voxel by connecting the voxel to its $p$-nearest neighboring voxels in a star topology. The voxel located at the center of a mesh is called the \textit{seed voxel}. If a voxel resides within the neighborhood of a seed voxel, then we connect it to the seed voxel with an edge, directed towards the seed voxel. As the name implies, the spatially local meshes are constructed by using the $p$-spatially nearest neighbors, whereas the functionally local  meshes are constructed by using the $p$-functional nearest neighbors of the seed voxels. Since we form the meshes around all of the voxels in the brain volume, the meshes defined over a neighborhood system may overlap. A seed voxel in a  mesh may become a neighboring voxel of a seed voxel in a different mesh. 

The formal definition of the local meshes are given below:

\textbf{Definition 3: Spatially Local Meshes (SLM):} For each voxel  $v_j$ in the brain volume,  a spatially local mesh, formed around that voxel, called the seed voxel, is defined as a directed graph $\mathcal{M}^{spat}_j = (\mathcal{V}^{spat}_j, \mathcal{E}^{spat}_{jk})$ where $\mathcal{V}^{spat}_j=\{v_j \cup \eta_{p}^{spat}[v_j]  \}$ represents a set of the nodes of the mesh, and 
\[
\mathcal{E}^{spat}_{j} = \{e_{jk} : \forall v_k \in \eta_{p}^{spat}[v_j]  \}
\]
represents the edges formed between the seed voxel $v_j$ of the mesh $\mathcal{M}^{spat}_j$ and its $p$-spatially nearest neighbors. The direction of the edges is taken towards the seed voxel (see Fig.~\ref{fig:lmm-tm}).

\textbf{Definition 4: Functionally Local Meshes (FLM):} 
For each voxel  $v_j$ in the brain volume,  a functionally local mesh which is constructed around that voxel is defined as a directed graph $\mathcal{M}^{func}_j = (\mathcal{V}^{func}_j, \mathcal{E}^{func}_{jk})$ where $\mathcal{V}^{func}_j=\{v_j \cup \eta_{p}^{func}[v_j]  \}$ represents a set of the nodes of the mesh, and 
\[\mathcal{E}^{func}_{j} = \{e_{jk} : \forall v_k \in \eta_{p}^{func}[v_j]  \}
\]
represents the edges formed between $v_j$ and its $p$-functionally nearest neighbors (see Fig. \ref{fig:fmm-tm}).

Notice that, for both neighborhood systems, a seed voxel located at the center of a mesh becomes a neighboring voxel in another mesh.  Therefore, there may be two directed edges between two neighboring voxels, $v_j$ and $v_k$, in opposite directions. The edge weights in each mesh are estimated by minimizing the mean square error of a linear model, as will be explained in the next section. In the proposed linear model, the seed voxel is represented in terms of linear combination of its $p$-nearest neighbors. The same star topology is defined for all meshes and employed in the experiments with a fixed mesh size $p$. In other words, the voxels and the edges formed between them in a mesh do not change in time. However, the weights of nodes which correspond to the intensity values measured for each entry of the time-series representing the BOLD response, change in time. Since we estimate a set of edge weights for each cognitive stimulus, the edges remain the same for the duration of a stimulus and only changes across the stimuli.

The size of each mesh is defined by the order of the neighborhood system, $p$. As $p$ gets larger, the size of a mesh constructed around each voxel increases, resulting in more and more overlaps among the meshes. Therefore, $p$ can be considered as a measure of the degree of locality of the proposed mesh model where we represent the BOLD response of a seed voxel as the linear combination of the BOLD responses of the neighboring voxels of the seed voxel. 

Once we define a mesh around each voxel, we address the problem of estimating the edge weights. The proposed edge estimation method is explained next.

\subsection{Estimation of Edge Weights of Meshes}

 Suppose that we record the BOLD response at each voxel $v_j$ that is located at coordinate $ \bar{l_j} $ during a stimulus $s_i$ in order to measure the brain activation for a predefined cognitive state with label  $c$. We denote each intensity value of the BOLD response measured at a voxel $v_j$ at time instance $ t_d $ for a stimulus $s_i$ as $v(s_i, t_d, \bar{l_j})$. For each stimulus presentation, we record $D$ measurements at each voxel, where $d = 1, 2, \ldots, D$.
 
For a given stimulus $s_i$, let us denote weight of an edge $e_{j,k}$ as  $w_i(e_{jk})$.  We approximate the weight of an edge   $w_i(e_{jk})$ directed from a voxel $v_k$ to $v_j$ for stimulus $s_i$ by estimating the value $ a_{i,j,k}$  when $v_j$ is considered as the seed voxel of a mesh.

We estimate the edge weights for both spatially and functionally local meshes in the same way. First, we obtain a BOLD response vector $\bar{r}(s_i, \bar{l_j}) \in \mathbb{R}^D$ by concatenating the voxel intensity values  $ v(s_i, t_d, \bar{l_j})$, for all time instances that are  recorded during a single stimulus, $s_i$,  such that:
\begin{equation}
\label{eq:response_from_voxel}
\bar{r}(s_i, \bar{l_j})= [ v(s_i, t_1, \bar{l_j}), v(s_i, t_2, \bar{l_j}), \ldots, v(s_i, t_D, \bar{l_j})]^T,
\end{equation}
where $d = 1, 2, \ldots, D$. When we concatenate the BOLD responses of all the stimuli, we obtain a vector of BOLD responses measured from a voxel $v_j$ for all training samples as follows; 
\begin{align}
\label{eq:all_response_from_voxel}
R(v_j)= [\bar{r}(s_1, \bar{l_j}), \bar{r}(s_2, \bar{l_j}), \ldots, \bar{r}(s_{N^{tr}}, \bar{l_j})], 
\end{align}
where $N^{tr}$ denotes the number of training samples. Notice that $R(v_j)$ is used to find the $p$-functionally nearest neighbors (see \eqref{eq:correlation}).

 We estimate the edge weights of a mesh formed around a voxel $v_j$ located at coordinate $l_j$ for a stimulus $s_i$ by a linear model formed among the BOLD response of a seed voxel of the mesh $\bar{r}(s_i, \bar{l_j})$, and the BOLD responses of its $p$-nearest neighbors  $\{\bar{r}(s_i, \bar{l_k}): v_k \in \eta_p[v_j]\}$ as follows;
\begin{equation}
	\label{eq:space_time_mesh}
	\bar{r}({s_i} ,{\bar{l}_j}) = \sum_{v_k \in \eta_p[v_j]}{ {a_{i,j,k}} \ \bar{r}(s_i,\bar{l}_k) + \bar{\varepsilon}_{i,j}} ,
\end{equation}
where $\bar{\varepsilon}_{i,j}$ is an error vector defined as 
\begin{align}
	\bar{\varepsilon}_{i,j}= (\varepsilon_{i,1,j}, \varepsilon_{i,2,j}, \ldots, \varepsilon_{i,K,j})^T.
\end{align}

Notice that $\eta_p$ corresponds to $\eta_p^{spat}$ if BOLD meshes are formed considering spatial neighborhood, and corresponds to $\eta_p^{func}$ if the meshes are formed considering functional neighborhood.

In \eqref{eq:space_time_mesh}, a BOLD response obtained from a seed voxel, $\bar{r}({s_i} ,{\bar{l}_j})$, is modeled using a weighted linear combination of the BOLD responses obtained from its $p$-nearest neighbors. A weight $a_{i,j,k}$ represents the relationship between responses obtained from a seed voxel $v_j$ and a neighboring voxel at $v_k$. In addition, each entry $v(s_i, t_d, \bar{l_j})$  of $\bar{r}(s_i, \bar{l_j})$ is represented by the same weighted combination of its neighboring entries $v(s_i, t_d, \bar{l_k})$ of $\bar{r}(s_i, \bar{l_k})$ in \eqref{eq:space_time_mesh}. Therefore, \eqref{eq:space_time_mesh} can be equivalently represented as follows: 

	\begin{equation}
	\label{eq:space_time_mesh_open}
	\begin{pmatrix}
	v(s_i, t_1, \bar{l_j})\\
	v(s_i, t_2, \bar{l_j})\\
	\vdots \\
	v(s_i, t_D, \bar{l_j})
	\end{pmatrix}= 
	\sum_{\bar{l}_b \in \eta_p}{a_{i,j,k}}
	\begin{pmatrix}
	v(s_i, t_1, \bar{l_k})\\
	v(s_i, t_2, \bar{l_k})\\
	\vdots \\
	v(s_i, t_D, \bar{l_k})
	\end{pmatrix}
	+ 
	\begin{pmatrix}
	\varepsilon_{i,1,j}\\
	\varepsilon_{i,2,j}\\
	\vdots \\
	\varepsilon_{i,D,j}
	\end{pmatrix}.
	\end{equation}

  The weights $ a_{i,j,k}$ of a mesh edge  $e_{j,k}$,  $\forall v_k \in \eta_p[v_j]$, are estimated by minimizing the expected square error defined as
\begin{equation}
	\label{eq:expected_value}
	E \left( \left( \bar{\varepsilon}_{i,j} \right) ^2 \right) = E \left( \left( \bar{r}({s_i} ,{\bar{l}_j}) - \sum_{v_k \in \eta_p} {a_{i,j,k}} \ \bar{r}(s_i,\bar{l}_k) \right) ^2 \right),
\end{equation}
where $E(\cdot)$ is the expectation operator applied over the time elapse of a BOLD response corresponding to a stimulus period $D$. We compute an edge vector ${\bar{a}_{i,j} = [a_{i, j, 1}, a_{i, j, 2}, \ldots, a_{i, j, p}]}$ using ridge regression as
\begin{equation}
	\label{eq:ridge_equation}
	\bar{a}_{i,j} = (Q_{i,j}^T Q_{i,j} + \lambda I)^{-1}Q_{i,j}^T\bar{r}({s_i} ,{\bar{l}_j}),
\end{equation}
where  $\lambda \in \mathbb{R}$ is a regularization parameter which is estimated during the training phase, and  $Q_{i,j}$ is a $ D \times p$ matrix consisting of BOLD responses obtained from $p$-nearest neighbors of a seed voxel $v_j$ during the presentation of sample $s_i$ such that
\begin{equation}
	\label{eq:R matrix}
	Q_{i,j} = [ \bar{r}({s_i} ,{\bar{l}_k}) ], \; \forall v_k \in \eta_p[v_j].
\end{equation}

One of the crucial tasks of forming local meshes is the identification of the degree of locality for each mesh. The number of the nearest neighbors, $p$, defines the size of each mesh. 
\begin{itemize}

\item For a spatially local mesh, when $p$ is small, the seed voxel of the mesh is related only to a few very spatially close voxels. As we increase $p$, the mesh includes larger areas in the brain volume to take into account the contribution of distant voxels to the seed. 

\item On the other hand, for a functionally local BOLD mesh, small $p$ implies that only a few most correlated voxels are related to the seed voxel. As we increase the mesh size, we include the contribution of the voxels that are relatively less correlated to the seed voxel.
\end{itemize}

Identification of the ``optimal" mesh size, $p$, is a crucial issue. In this study we optimize the mesh size using a validation set. This task can also be achieved by using an  Information Theoretic function, such as Akaike's information criterion (AIC) \cite{Akaike1973}, Bayesian Information Criterion (BIC) \cite{Schwarz1978} or Rissannen's Minimum Description Length (MDL) \cite{Rissanen1978} to estimate the model order. It is possible to define a different mesh size for each seed voxel by using one of the approaches mentioned above. The interested reader is referred to \cite{Onal2014}. In this study, we select a fixed mesh size for the entire brain volume, for each stimulus.

\section{Classification of Cognitive States For Brain Decoding}

The representation power of the proposed mesh ensemble is analyzed in the fMRI recordings during brain encoding and decoding tasks. For this purpose, we employ machine learning methods where encoding of a cognitive state is represented by the training phase and the decoding state corresponds to the recognition phase of a classification algorithm. The major question is then how well a cognitive state can be represented by the ensemble of meshes? The answer to this question can be partly observed from the performance of the classifier. 

 Recall that for each cognitive stimulus $s_i$ with a label $c$, we record a sequence of brain volumes, where the number of brain volumes  $D$, in this sequence depends on the duration of the stimulus. This sequence of brain volumes is considered as one sample with a label $c$, in the training set.  In order to represent a cognitive state corresponding to a stimulus $s_i$ by the mesh ensembles,  we estimate the edge weights of all local meshes extracted from the corresponding sequence of the brain volumes. The sequence of brain volumes recorded during the stimulus $s_i$ is then represented by a brain graph, which consists of the ensemble of local meshes. Finally, each sample in the dataset is represented by a vector with the entries of the estimated  mesh arc weights in the input space of a classifier.  

Formally speaking, for each sample $s_i$ with label $c$, the edge weights $\bar{a}_{i,j} \in \mathbb{R}^{1 \times p}$ are used to construct a feature vector ${F_i = [ \bar{a}_{i,1}, \bar{a}_{i,2}, \ldots, \bar{a}_{i,M} ] }$ by concatenating all edge vectors $\bar{a}_{i, j}$, $ \forall j=1,2,\ldots,M$, where $M$ is the number of voxels. We denote a feature vector of a sample $s_i$ belonging to training set as $F^{tr}_i$, while $F^{te}_i$ is a feature vector of a sample $s'_i$ belonging to test set. Then, we construct a set of features of training samples $S^{tr} =\{ F^{tr}_i \}_{i=1}^{N^{tr}}$ and a set of features of test samples $S^{te} =\{ F^{te}_i \}_{i=1}^{N^{te}}$ to train and test the classifiers for classification of cognitive states.

%\begin{figure}
%	\centering
%	%\shorthandoff{=}

%	\includegraphics[scale=0.35]{experiment_memory.png}
%	%\shorthandon{=}
%	\caption{A sample trial from memory retrieval fMRI experiment. In each trial participant is presented a study list belonging to one of ten categories (animal category is shown). Each trial starts with the presentation of study list for 2 seconds in encoding phase, followed by a delay period of mathematical problem solving and ends with a test probe in the retrieval phase. The participant is expected to recognize whether the word was a member of the current study list or not \cite{Oztekin2011}.}
%	\label{fig:experiment_memory}
%\end{figure}

\section{Brain Decoding Experiments Performed on the fMRI Datasets}

In order to observe the validity of the proposed mesh models, we perform two groups of experiments. In the first group, we train and test a Support Vector Machine (SVM) classifier by using the labeled fMRI data recorded using the experimental setups explained in Section II. In this group of experiments, we examine the power of the models for representing brain encoding and decoding tasks. In the second group of experiments, we analyze the validity of the local meshes by exploring the similarities among the voxel time series in each mesh, and the distribution of the error of the proposed linear regression model. 

\subsection{Comparison of the Spatial and Functional Mesh Models with {State-of-the-Art} Methods}

 In order to compare the proposed spatial and functional meshes with the state-of-the-art MVPA methods, we measure the encoding and decoding performances of the models.  The encoding process is performed by training an SVM classifier with linear kernel, and decoding process corresponds to classification of an unknown stimulus with class label $c$. We perform seven sets of computer experiments on two different groups of fMRI datasets:
 \begin{itemize}

\item In the first and second set of experiments, we represent a cognitive stimulus using the spatial and functional meshes. The classification performances are measured by using the edge weights obtained using the spatially  \textbf{(SLM)} and functionaly local mesh \textbf{(FLM)} models. 
 
\item In the third and fourth set of experiments, we  tested the classification performances by employing Local Mesh Model (\textbf{LMM}) and Functional Mesh Model(\textbf{FMM}). Recall that both \textbf{LMM} and \textbf{FMM} keep only a single value from the BOLD response for each voxel omitting the rest of the signals measured during a stimulus. 

\item In the fifth group of experiments, we test the performance of SVM  classifiers which employ features that are computed using pairwise Pearson correlation (\textbf{FC-mesh}). First, we compute the pairwise correlations between the voxel BOLD responses given by seed voxels and each of their $p$-nearest neighbors for each stimulus. In other words, first meshes are constructed around seed voxels, and then pairwise correlations are computed between voxel pairs within meshes instead of estimating the edge weights within a neighborhood. Then, we form our feature vector by concatenating the correlation values within all meshes. Note that, we obtain feature vectors whose sizes are equal to that of \textbf{SLM} and \textbf{FLM} by concatenating only the distances within meshes to make a fair comparison.

%Therefore, the feature vectors has nearly 3.2 million elements. 
%Employing such high dimensional data during classification would lead to a curse of dimensionality problem.

\item In the sixth set of experiments, we analyze the classification performance of the classifiers for the state-of-the-art MVPA methods in which raw voxel intensity values are used as features to train and test classifiers. For \textbf{MVPA-peak}, we only used the third instance $v(s_i, t_3, \bar{l_j})$ of a BOLD response $\bar{r}({s_i} ,{\bar{l}_j})$ following our assumption that if a voxel becomes active, then its HRF reaches to its peak value after 5-6 seconds (around $t_3$). On the other hand, \textbf{MVPA-mean} depicts the case where we employed the average of $d$ measurements of $\bar{r}({s_i} ,{\bar{l}_j})$. We also concatenate each of the $d$ measurements of BOLD response  $\{v(s_i, t_d, \bar{l_j})\}_{d=1}^D$ to obtain results for \textbf{MVPA-all}. The dimension of a feature vector constructed for \textbf{MVPA-peak} and \textbf{MVPA-mean} equals to 1254 for visual object recognition experiment, and 3000 for emotional memory retrieval experiment. Moreover, since we concatenate all measurements for \textbf{MVPA-all}, the size of the feature vectors used for \textbf{MVPA-all} equals to $d$ times the size of features vectors for \textbf{MVPA-mean} and \textbf{MVPA-peak}. 

\item In the seventh set of experiments, we selected the surrounding voxels in the mesh as the random voxels to test the importance of locality. In \textbf{LM-rand}, we form meshes defined over a sequence of brain volumes with random voxels.

 \end{itemize}

\renewcommand{\arraystretch}{1.2}
\begin{table}[t]
	\centering
	\caption{Classification performance (\%) of SVM classifier computed in visual object recognition experiment.}
	\label{tab:object_results}
	\begin{tabular}{|l|c|c|c|c|c|c|}
		\hline
		& $\boldsymbol{P_1}$ & $\boldsymbol{P_2}$ & $\boldsymbol{P_3}$ & $\boldsymbol{P_4}$ & $\boldsymbol{P_5}$ & $\boldsymbol{Avg}$ \\ \hline
		\textbf{FLM}    & 93(16)                 & 83(5)                 & 89(16)                 & 81(17)                 & 61(4)                & 81.4                 \\ \hline
		\textbf{SLM}    & 80(5)                 & 81(30)                 & 81(14)                 & 72(10)                 & 69(6)                 & 76.6                 \\ \hline
		\textbf{FMM-mean}  & 77(19)                & 67(27)                 & 72(23)                 & 75(18)                 & 64(15)                 & 71.0                 \\ \hline
		\textbf{FMM-peak}  & 63(26)                 & 69(23)                 & 72(14)                 & 75(30)                 & 67(11)                 & 69.2                 \\ \hline
		\textbf{LMM-mean}  & 73(21)                 & 67(27)                 & 69(30)                 & 69(25)                 & 64(22)                 & 68.4                 \\ \hline
		\textbf{LMM-peak}  & 73(23)                 & 72(28)                 & 78(29)                 & 58(19)                 & 69(10)                 & 70.0                 \\ \hline
		\textbf{LM-rand}   & 74                 & 81                 & 81                 & 72                 & 69                 & 75.4                 \\ \hline
		\textbf{FC-mesh}   & 65                 & 74                 & 61                 & 58                 & 50                 & 61.8                 \\ \hline
		\textbf{MVPA-mean} & 70                 & 71                 & 72                 & 78                 & 69                 & 72.0                 \\ \hline
		\textbf{MVPA-peak} & 68                 & 67                 & 69                 & 83                 & 72                 & 71.8                 \\ \hline
		\textbf{MVPA-all} & 68                 & 76                 & 78                 & 69                 & 68                 & 71.8                 \\ \hline
	\end{tabular}
\end{table}

\renewcommand{\arraystretch}{1.2}
\begin{table*}[t]
	\centering
	\caption{Classification performance (\%) of SVM classifier computed in emotional memory retrieval experiment (2-class).}
	\label{tab:emotion_results}
	\begin{tabular}{|l|c|c|c|c|c|c|c|c|c|c|c|c|c|c|}
		\hline
		& $\boldsymbol{P_1}$ & $\boldsymbol{P_2}$ & $\boldsymbol{P_3}$ & $\boldsymbol{P_4}$ & $\boldsymbol{P_5}$ & $\boldsymbol{P_6}$ & $\boldsymbol{P_7}$ & $\boldsymbol{P_8}$ & $\boldsymbol{P_9}$ & $\boldsymbol{P_{10}}$ & $\boldsymbol{P_{11}}$ & $\boldsymbol{P_{12}}$ & $\boldsymbol{P_{13}}$ & $\boldsymbol{Avg}$ \\ \hline
		\textbf{FLM}    & 71(12)                 & 75(5)                 & 73(7)                 & 82(12)                 & 81(8)                 & 79(5)                 & 88(9)                 & 78(10)                 & 81(13)                 & 79(9)                  & 72(5)                  & 84(14)                  & 66(8)                  & 77.6               \\ \hline
		\textbf{SLM}    & 72(5)                 & 75(5)                 & 78(12)                 & 75(6)                 & 78(5)                 & 74(6)                 & 88(5)                 & 78(11)                 & 81(11)                 & 82(14)                  & 81(15)                  & 70(11)                  & 66(15)                  & 76.8               \\ \hline
		\textbf{FMM-mean}  & 50(8)                 & 61(5)                 & 66(5)                 & 63(5)                 & 62(10)                 & 64(5)                 & 82(9)                 & 63(6)                 & 75(7)                 & 72(5)                  & 60(5)                  & 63(5)                  & 54(5)                  & 64.2               \\ \hline
		\textbf{FMM-peak}  & 66(15)                 & 67(9)                 & 69(9)                 & 78(14)                 & 72(12)                 & 63(11)                 & 76(6)                 & 69(6)                 & 75(14)                 & 68(5)                  & 75(15)                  & 91(14)                  & 60(11)                  & 71.5               \\ \hline
		\textbf{LMM-mean}  & 51(7)                 & 60(7)                 & 62(5)                 & 70(9)                 & 69(12)                 & 54(7)                 & 86(8)                 & 66(7)                 & 74(9)                 & 70(5)                  & 58(5)                  & 58(7)                  & 70(7)                  & 65.2               \\ \hline
		\textbf{LMM-peak}  & 58(12)                 & 65(7)                 & 62(8)                 & 81(15)                 & 70(15)                 & 69(7)                 & 83(14)                 & 66(10)                 & 74(13)                 & 68(5)                  & 71(9)                  & 83(14)                  & 76(14)                  & 71.2               \\ \hline
		\textbf{LM-rand}   & 53                  & 63                  & 65                  & 65                  & 60                  & 50                  & 77                  & 59                  & 74                  & 64                   & 53                   & 64                   & 54                   &  61.6                  \\ \hline
		
		\textbf{FC-mesh}   & 55                  & 53                  & 57                  & 59                  & 56                  & 56                  & 58                  & 63                  & 57                  & 57                   & 62                   & 67                   & 56                   &  55.7                  \\ \hline
		\textbf{MVPA-mean} & 57                 & 58                 & 68                 & 63                 & 61                 & 50                 & 81                 & 60                 & 76                 & 47                  & 55                  & 54                  & 61                  & 60.8               \\ \hline
		\textbf{MVPA-peak} & 72                 & 66                 & 67                 & 77                 & 70                 & 69                 & 83                 & 70                 & 82                 & 67                  & 78                  & 81                  & 73                  & 73.4               \\ \hline
		\textbf{MVPA-all} & 66                 & 61                 & 60                 & 69                 & 59                 & 73                 & 80                 & 65                 & 74                 & 72                  & 72                  & 63                  & 71                  & 68.1               \\ \hline
	\end{tabular}
\end{table*}

\renewcommand{\arraystretch}{1.2}
\begin{table*}[t]
	\centering
	\caption{Classification performance (\%) of SVM classifier computed in emotional memory retrieval experiment (4-class).}
	\label{tab:emotion_results_4class}
	\begin{tabular}{|l|c|c|c|c|c|c|c|c|c|c|c|c|c|c|}
		\hline
		& $\boldsymbol{P_1}$ & $\boldsymbol{P_2}$ & $\boldsymbol{P_3}$ & $\boldsymbol{P_4}$ & $\boldsymbol{P_5}$ & $\boldsymbol{P_6}$ & $\boldsymbol{P_7}$ & $\boldsymbol{P_8}$ & $\boldsymbol{P_9}$ & $\boldsymbol{P_{10}}$ & $\boldsymbol{P_{11}}$ & $\boldsymbol{P_{12}}$ & $\boldsymbol{P_{13}}$ & $\boldsymbol{Avg}$ \\ \hline
		\textbf{FLM}    & 59(6)                 & 54(10)                 & 62(6)                 & 68(13)                 & 60(6)                 & 66(6)                 & 75(9)                 & 59(15)                 & 67(5)                 & 74(13)                  & 64(7)                  & 62(14)                  & 58(5)                  & 63.7               \\ \hline
		\textbf{SLM}    & 58(13)                 & 53(13)                 & 63(8)                 & 67(14)                 & 63(8)                 & 62(5)                 & 78(11)                 & 62(5)                 & 75(5)                 & 74(9)                  & 58(9)                  & 58(12)                  & 57(12)                  & 63.7               \\ \hline
		\textbf{FMM-mean}  & 38(5)                 & 42(8)                 & 46(14)                 & 44(14)                 & 49(10)                 & 42(5)                 & 74(15)                 & 36(5)                 & 57(7)                 & 52(6)                  & 42(9)                  & 32(5)                  & 35(7)                  & 45.1               \\ \hline
		\textbf{FMM-peak}  & 42(10)                 & 46(6)                 & 44(9)                 & 54(15)                 & 40(14)                 & 46(5)                 & 58(7)                 & 37(12)                 & 58(11)                 & 57(7)                  & 58(13)                  & 43(14)                  & 45(13)                  & 48.3               \\ \hline
		\textbf{LMM-mean}  & 38(11)                 & 46(6)                 & 50(8)                 & 47(12)                 & 47(14)                 & 42(5)                 & 68(12)                 & 42(6)                 & 51(8)                 & 52(7)                  & 38(7)                  & 35(5)                  & 38(6)                  & 45.7               \\ \hline
		\textbf{LMM-peak}  & 42(9)                 & 36(12)                 & 44(12)                 & 45(11)                 & 47(15)                 & 48(7)                 & 58(11)                 & 32(8)                 & 61(15)                 & 57(7)                  & 47(10)                  & 46(11)                  & 47(10)                  &  47.0               \\ \hline
		\textbf{LM-rand}   & 37                  & 43                  & 37                  & 42                  & 40                  & 40                  & 63                  & 44                  & 59                  & 51                   & 39                   & 45                   & 39                   & 44.6                   \\ \hline	
		\textbf{FC-mesh}   & 31                  & 35                  & 26                  & 33                  & 32                  & 30                  & 36                  & 34                  & 31                  & 39                   & 38                   & 41                   & 33                   & 33.7                   \\ \hline
		\textbf{MVPA-mean} & 37                 & 40                 & 44                 & 45                 & 45                 & 37                 & 67                 & 41                 & 58                 & 40                  & 40                  & 33                  & 34                  & 43.2               \\ \hline
		\textbf{MVPA-peak} & 49                 & 45                 & 44                 & 55                 & 41                 & 51                 & 64                 & 41                 & 57                 & 49                  & 57                  & 52                  & 46                  & 50.1               \\ \hline
		\textbf{MVPA-all} & 43                 & 41                 & 40                 & 44                 & 34                 & 43                 & 57                 & 39                 & 53                 & 41                  & 35                  & 38                  & 42                  & 42.4               \\ \hline
	\end{tabular}
\end{table*}

%Finally, we applied principal component analysis (PCA) and transformed our data onto first 200 components, which is determined experimentally. Afterwards, we computed two more experiments to understand whether our method is also valid for transformed data. In the first experiment, we directly used this data as features and applied classical MVPA method. In the second experiment we formed a mesh using the transformed data and computed the relationships among components. Since the location information about principal components does not exist, we computed the Pearson correlation of all pair-wise component values and form the mesh with functionally nearest neighbors as in \cite{Firat2013}. Extracted edge weights $a_{i,j,k}$ are used as features to classification. In other words, we applied STMM using transformed data onto principal components.

\subsection{Classification Results for Brain Decoding}

We perform intra-subject classification using SVM classifiers, and features that are extracted from samples collected in the proposed visual object recognition experiment for five participants, and emotional memory retrieval experiment for thirteen participants. The same mesh size is used for each mesh that was computed using the data collected for the same participant.

\textbf{Visual Object Recognition Experiment:}
Table \ref{tab:object_results} provides the classification performance of SVM using the models described in the previous subsection. The regularization parameter $\lambda$ is selected experimentally as $\lambda = 0.5$. We computed meshes using various mesh sizes ${p \in \mathcal{P} = \{2,3,\ldots,30\} }$. In order to optimize the mesh size within this interval, we randomly split a part of test data as our validation set. We ensured that our validation and test sets are balanced. We selected the optimal mesh sizes as the ones that maximizes the validation performances. 

Classification results of visual object recognition experiment are given in Table \ref{tab:object_results}. Additionally, the performances obtained on test set using the \textit{optimal} mesh size $\hat{p}$ obtained from validation sets are given in Table \ref{tab:object_results} for each method employed by each classifier and for each participant. For example, we obtain $93\%$ accuracy for \textbf{FLM} when the mesh size $\hat{p}$ is $16$ for the first participant $P_1$. It can be observed that, the mesh size $\hat{p}$ which leads to the maximum performance varies for different participants and methods. In addition, classifiers which employ the features extracted for \textbf{FLM} and \textbf{SLM} perform the best among others, where the features extracted for \textbf{FLM}  perform slightly better than the features extracted for \textbf{SLM}. Note that, selecting random voxels in \textbf{LM-rand} performs worse than employing spatial or functional neighbors.

\begin{figure}[t!]
	\centering
	%\shorthandoff{=}
	\includegraphics[trim=4cm 4cm 4cm 4cm, scale=0.35]{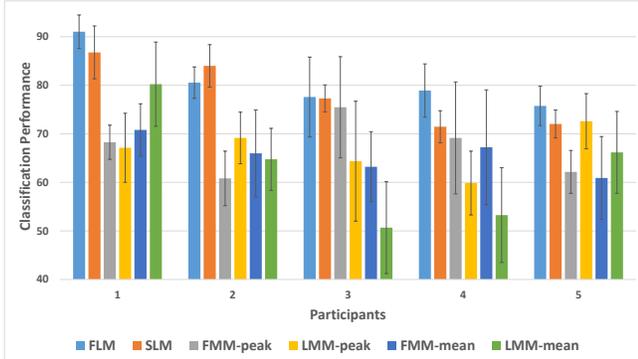}
	%\shorthandon{=}
	\caption{Mean and standard deviation of classification performances of SVM compute din visual object recognition experiment.}
	\label{fig:robustness_object}
\end{figure}

\begin{figure*}[t]
	\centering
	%\shorthandoff{=}
	\includegraphics[trim=4cm 4cm 4cm 4cm, scale=0.7]{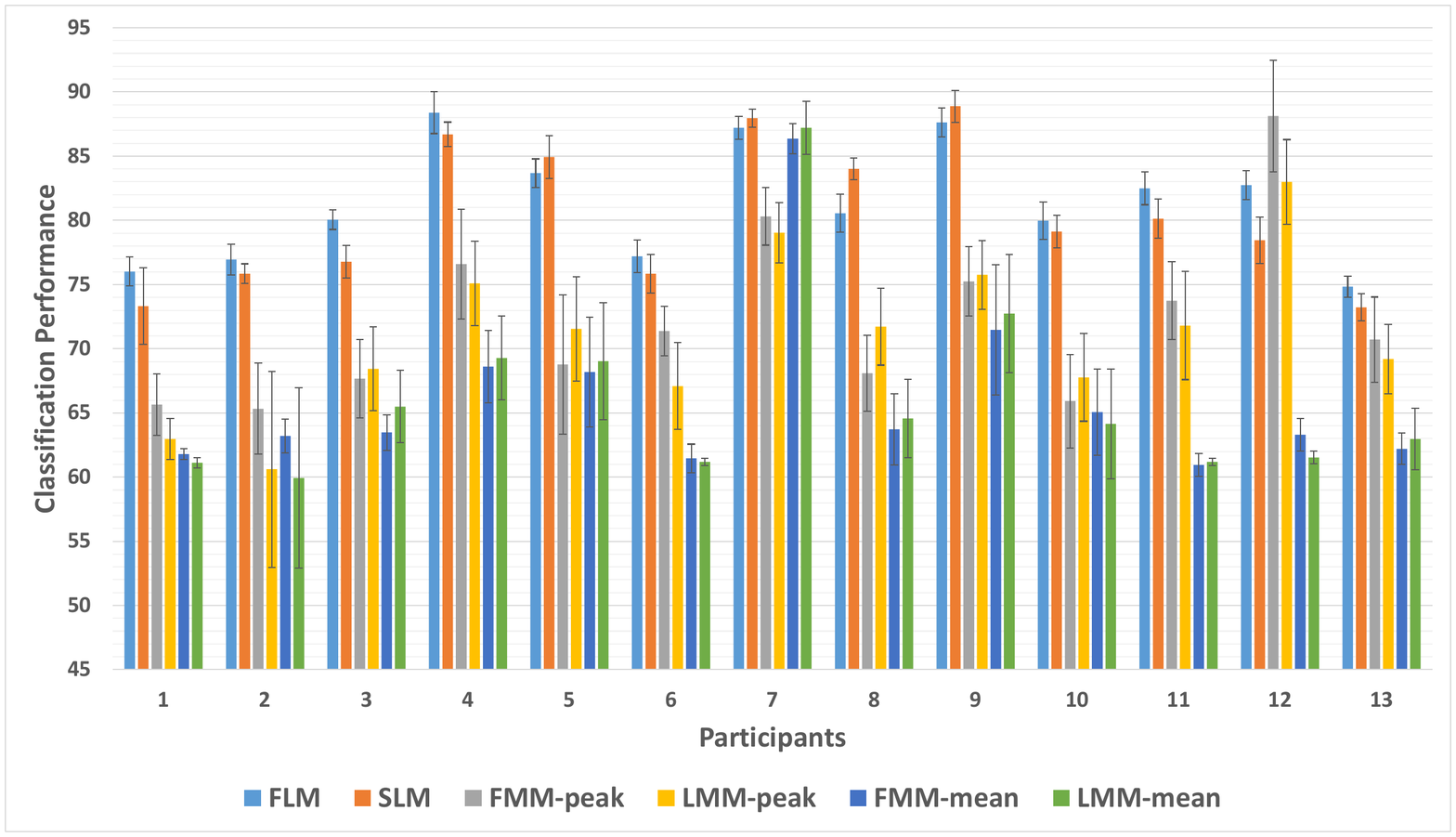}
	%\shorthandon{=}
	\caption{Mean and standard deviation of classification performances of SVM computed in emotional memory retrieval experiment (2-class).}
	\label{fig:robustness_emotional}
\end{figure*}

\begin{figure*}[t!]
	\centering
	%\shorthandoff{=}
	\includegraphics[trim=4cm 4cm 4cm 4cm, scale=0.7]{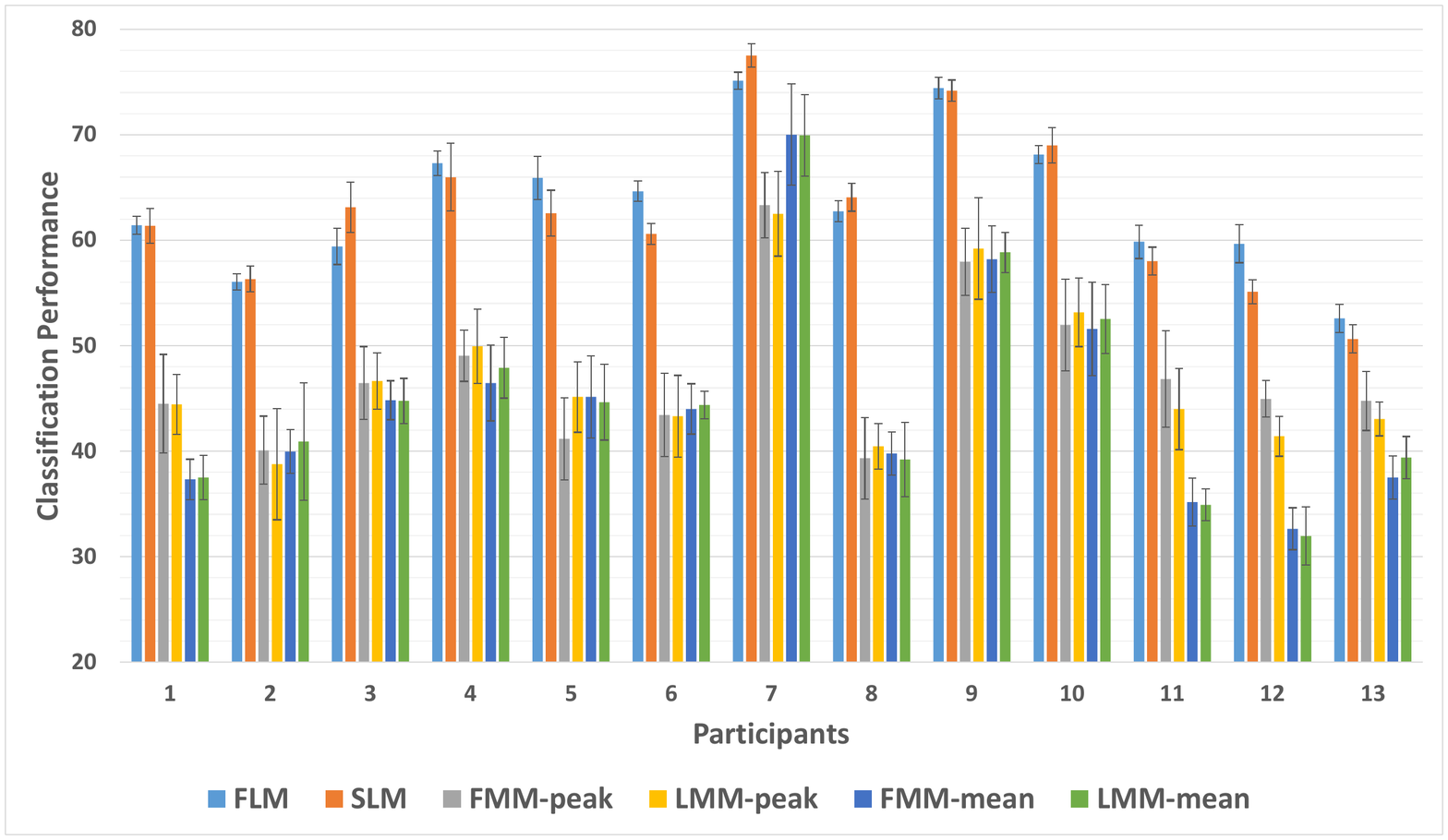}
	%\shorthandon{=}
	\caption{Mean and standard deviation of classification performances of SVM computed in emotional memory retrieval experiment (4-class).}
	\label{fig:robustness_emotional_4class}
\end{figure*}

\begin{figure*}[t!]
	\centering
	\begin{subfigure}[t]{0.3\textwidth}
		\includegraphics[width=\textwidth]{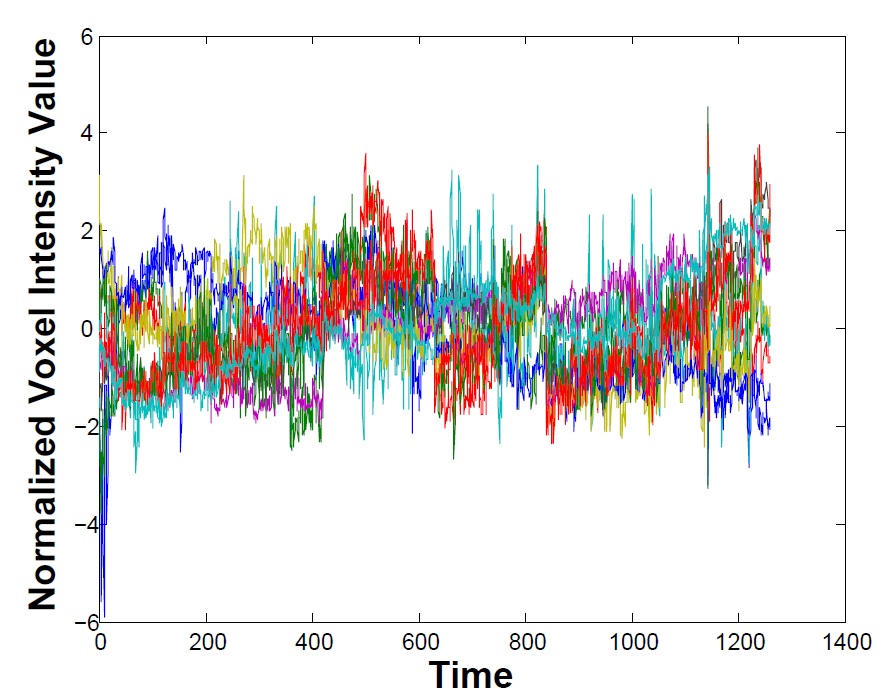}
		\caption{Random voxels.}
		\label{fig:random}
	\end{subfigure}
	~ %add desired spacing between images, e. g. ~, \quad, \qquad, \hfill etc. 
	%(or a blank line to force the subfigure onto a new line)
	\begin{subfigure}[t]{0.3\textwidth}
		\includegraphics[width=\textwidth]{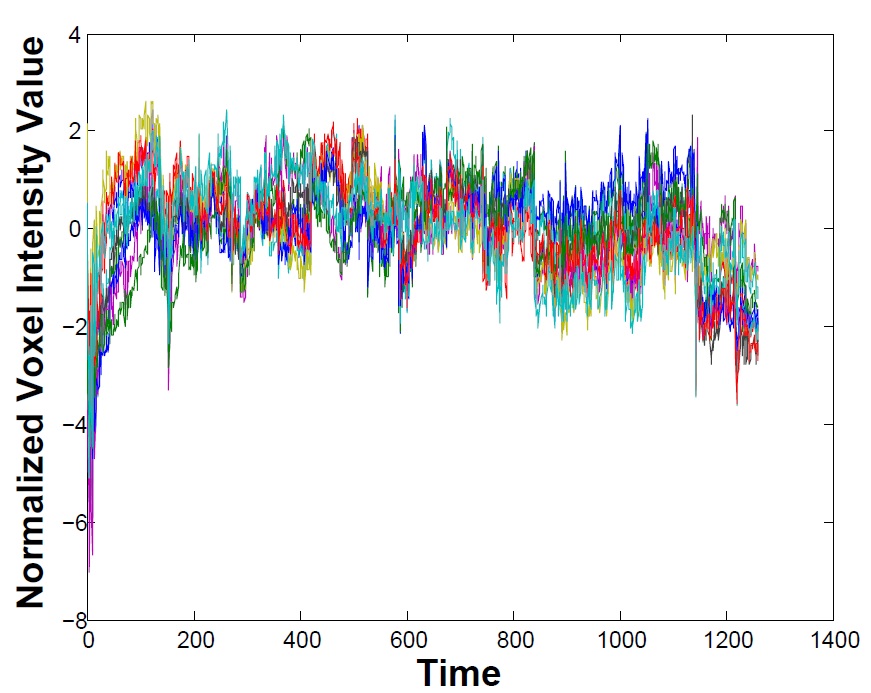}
		\caption{Spatial Neighbors.}
		\label{fig:spatial_neighs}
	\end{subfigure}
	~ %add desired spacing between images, e. g. ~, \quad, \qquad, \hfill etc. 
	%(or a blank line to force the subfigure onto a new line)
	\begin{subfigure}[t]{0.3\textwidth}
		\includegraphics[width=\textwidth]{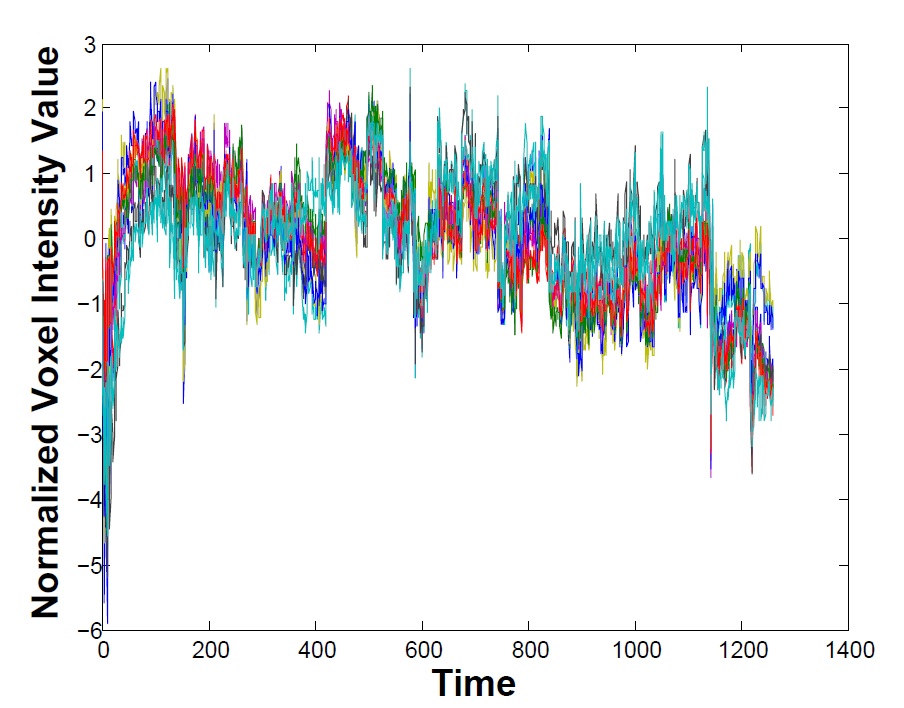}
		\caption{Functional Neighbors. }
		\label{fig:functional_neighs}
	\end{subfigure}
	\caption{BOLD responses of voxel sets.}
\end{figure*}

\begin{figure*}[b!]
	\centering
	\begin{subfigure}[t]{0.3\textwidth}
		\includegraphics[width=\textwidth]{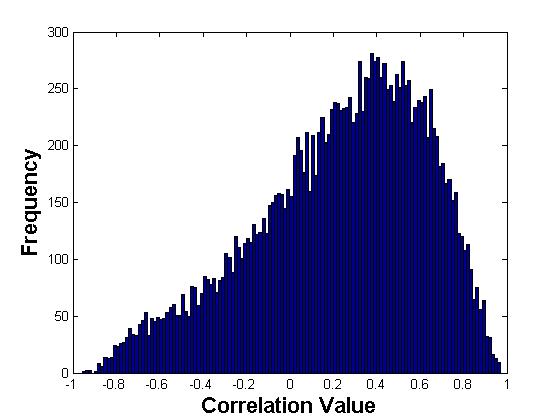}
		\caption{Random voxels.}
		\label{fig:random_corr}
	\end{subfigure}
	~ %add desired spacing between images, e. g. ~, \quad, \qquad, \hfill etc. 
	%(or a blank line to force the subfigure onto a new line)
	\begin{subfigure}[t]{0.3\textwidth}
		\includegraphics[width=\textwidth]{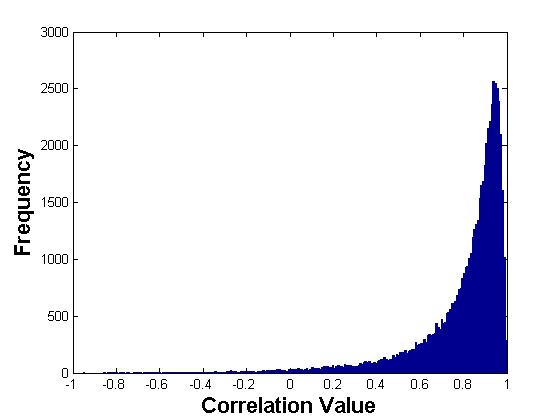}
		\caption{Spatial Neighbors.}
		\label{fig:spatial_corr}
	\end{subfigure}
	~ %add desired spacing between images, e. g. ~, \quad, \qquad, \hfill etc. 
	%(or a blank line to force the subfigure onto a new line)
	\begin{subfigure}[t]{0.3\textwidth}
		\includegraphics[width=\textwidth]{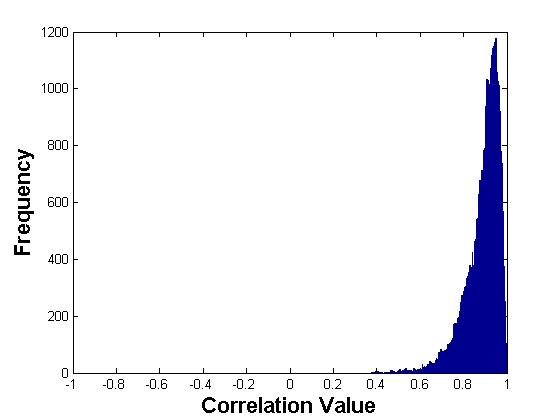}
		\caption{Functional Neighbors.}
		\label{fig:functional_corr}
	\end{subfigure}
	\caption{Histograms of correlations between seed voxels and surrounding voxels.}\label{fig:correlations}
\end{figure*}

\textbf{Emotional Memory Retrieval Experiment:} We provided results for 2-class and 4-class classification experiments. In these experiments, $\lambda$ is selected experimentally as $\lambda = 4$, and we computed meshes with mesh sizes ${p \in \mathcal{S} = \{5,6,\ldots,15\} }$. In the first case, we provided two-class classification results where classes correspond to neutral and emotional categories (see Table \ref{tab:emotion_results}). In the second case, four classes correspond to kitchen utensil, furniture, fear and disgust, where kitchen utensil and furniture belong to the neutral category, and fear and disgust belong to the emotional category (see Table \ref{tab:emotion_results_4class}). 

%We selected a random part of test set as validation set to optimize mesh sizes. (Bu ne demek?)

Similar to visual object recognition experiment, optimal mesh sizes are selected as the ones that maximize validation set accuracy within the interval $\mathcal{S}$. Table \ref{tab:emotion_results} and Table \ref{tab:emotion_results_4class} provide classification performance obtained using the optimal mesh sizes. The results of 2-class and 4-class classification experiments show that we obtain significantly better performance using features extracted for \textbf{FLM} and \textbf{SLM} than the other features. Employing random voxels in \textbf{LM-rand} performs significantly worse than employing spatial or functional neighbors. We have also observed that we obtain better performance using the features extracted by the employment of the peak values of BOLD responses than using the features extracted by computing the average of BOLD responses. In other words, we obtain better performance for \textbf{FMM-peak}, \textbf{LMM-peak} and \textbf{MVPA-peak} than for \textbf{FMM-mean}, \textbf{LMM-mean} and \textbf{MVPA-mean}, respectively.

First of all, the results show that employing a set of BOLD responses for the computation of meshes provides the most discriminate representations of cognitive states. Moreover, we observe that estimating functional relationships within a neighborhood provides better performance than using pairwise functional relationships as utilized for \textbf{FC-mesh}. Features extracted for \textbf{FC-mesh}  perform the worst among the others since computation of correlation among voxels using only a few voxel intensity values for each stimulus presentation is not enough to represent the real correlation among voxels computed considering the whole dataset. Therefore, we obtain better classification performance by computing a single functional connectivity matrix for all training data, and using the matrix to select functionally nearest neighbors, instead of computing the functional connectivity matrices for each sample and using their elements as features. Finally, we obtained worse classification performance using the classifiers which employ \textbf{MVPA} features consisting of individual voxel intensity values compared to our proposed methods.

We have provided the mean and standard deviations of classification performances using different meshes that are constructed with $p \in \mathcal{P}$ for visual object recognition experiment in Fig. \ref{fig:robustness_object}, and with $p \in \mathcal{S}$ for emotional memory retrieval experiment in Fig. \ref{fig:robustness_emotional} and Fig. \ref{fig:robustness_emotional_4class}. In Fig. \ref{fig:robustness_object}, bars represent the mean classification performance averaged over all $p \in \mathcal{P}$, whereas in Fig. \ref{fig:robustness_emotional} and Fig. \ref{fig:robustness_emotional_4class}, bars  represent the mean classification performance averaged over all $p \in \mathcal{S}$. In these figures, each error bar represents standard deviations of the corresponding performance value. For each participant, the first two bars depict performances of the methods that employ functional and spatial meshes, whereas the last four bars depict performances obtained using mesh models which do not consider whole BOLD responses. We have observed that standard deviation of performance values obtained using various mesh sizes decreases when temporal measurements are considered for statistical learning of the models. In other words, the methods employing temporal measurements recorded within neighborhoods are more robust to changes of mesh size $p$ than the methods which do not employ temporal measurements.

\subsection{Analysis of Local Meshes}

In the proposed local mesh models, we assume that the BOLD responses of voxels that are located in a neighborhood are similar to each other so that a BOLD response of a voxel can be represented by a linear combination of the BOLD responses of its nearest neighbors. In other words, we assume that the error of a linear regression model is small enough such that a cognitive stimulus can be represented by a set of local meshes, where the edge weights of the meshes can be estimated by minimizing the expected square error. 

\subsubsection{Statistical Analysis of Correlations of Voxels}
In order to understand how voxels within a neighborhood behave, and how their behavior differs from the behavior of random voxel sets, we plot a seed voxel with $i)$ a set of randomly selected voxels (Fig. \ref{fig:random}), $ii)$ its ten spatially (Fig. \ref{fig:spatial_neighs}) and $iii)$ ten functionally (Fig. \ref{fig:functional_neighs}) nearest neighbors. We observe that, sample voxels located within spatial and functional neighborhoods perform similar under presentation of the same stimuli. 

In the analysis, Pearson correlation is used to compute pairwise relationship (i.e. correlation) between the seed voxels and the surrounding voxels. In our analysis, surrounding voxels are selected as (a) random voxels, (b) spatially nearest neighbors of seed voxels and (c) functionally nearest neighbors of seed voxels. First, we compute correlations between ten surrounding voxels and the seed voxel. Then, we depict the histograms of all correlations computed for all voxels with their surrounding voxels in meshes. 

We observe that if meshes are formed with random voxels, then the mean of distribution of correlations lie around 0.4 (see Fig. \ref{fig:random_corr}), in other words, the randomly selected voxels may not have a statistically significant linear relationship with each other. On the other hand,  the results show that correlation values of the voxels observed with the highest frequency are close to 0.9 when the histograms are computed for meshes formed using spatially (see Fig. \ref{fig:spatial_corr}) and functionally (Fig. \ref{fig:functional_corr}) close voxels. These observations indicate that the voxels modeled in the same mesh have better statistical relationship with each other than the randomly selected meshes. Therefore, one may expect that the linear relationship among the BOLD responses observed in voxels located in a proposed neighborhood system provides a reasonable fit to the data.

\subsubsection{Statistical Analysis of Models}
\begin{figure*}[t]
	\centering
	\begin{subfigure}[t]{0.3\textwidth}
		\includegraphics[width=\textwidth]{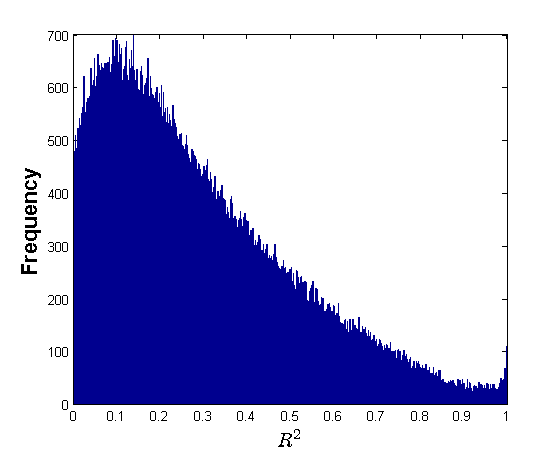}
		\caption{Random voxels.}
		\label{fig:random_rsquare}
	\end{subfigure}
	~ %add desired spacing between images, e. g. ~, \quad, \qquad, \hfill etc. 
	%(or a blank line to force the subfigure onto a new line)
	\begin{subfigure}[t]{0.3\textwidth}
		\includegraphics[width=\textwidth]{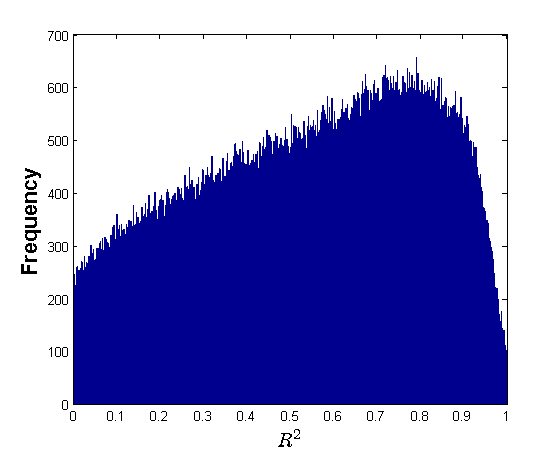}
		\caption{Spatial Neighbors.}
		\label{fig:spatial_rsquare}
	\end{subfigure}
	~ %add desired spacing between images, e. g. ~, \quad, \qquad, \hfill etc. 
	%(or a blank line to force the subfigure onto a new line)
	\begin{subfigure}[t]{0.3\textwidth}
		\includegraphics[width=\textwidth]{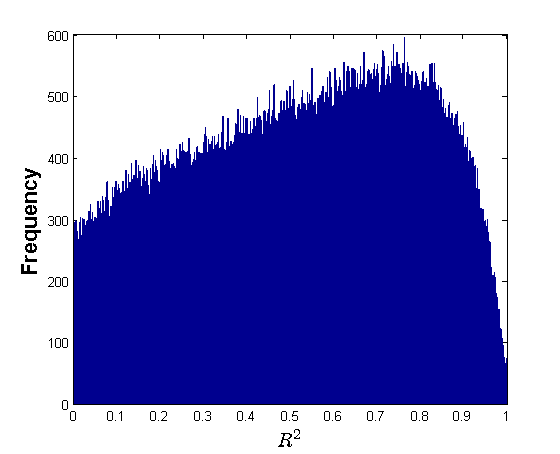}
		\caption{Functional Neighbors.}
		\label{fig:functional_rsquare}
	\end{subfigure}
	\caption{Histograms of $R^2$ values. }\label{fig:rsquares}
\end{figure*}

Recall that, we represent the voxel intensity values each voxel as a linear combination of those values of its $p$-nearest neighbors in our models. Yet, we obtain a regression error for each mesh for the estimation of seed voxels considering its neighbors. In the next set of experiments, we analyze how well the seed voxels are represented in terms of their nearest neighbors by exploring the model estimation error of the proposed  representation. 

For this purpose, we employ a goodness of fit measure called $R^2$ defined as one minus the residual variation divided by total variation, where residual variation ($SS^{r}$) is  the summation of square errors obtained from regression model, and total variation ($SS^{t}$) is the variance of the distribution of the actual data. 

More precisely, we compute the residual variation as,
\begin{equation}
	\label{eq:SSreg}
	SS^{r}_{i,j} = \sum_{d = 1} ^ {D} \varepsilon_{i,d,j}^2 ,
\end{equation}
where $\varepsilon_{i,d,j}$ denotes the error in \eqref{eq:space_time_mesh_open}, and $D$ represents the number of measurements recorded during each stimulus representation. On the other hand, total variance is computed as  
\begin{equation}
	\label{eq:SStot}
	SS^{t}_{i,j} = \sum_{d = 1} ^ {D} v(s_i, t_d, \bar{l_j})^2.
\end{equation}

Then, the $R^2$ is computed as
\begin{equation}
	\label{eq:Rsquared}
	R^2_{i,j} = 1 - \left( \frac{ SS^{r}_{i,j} }{ SS^{t}_{i,j}}\right) .
\end{equation}

Notice that the $R^2$ measure takes values between $0$ and $1$ such that the values closer to one represent better fit of models. In the analysis, we computed $R^2$ values for all meshes of a participant formed for all samples and around all voxels. In Fig. \ref{fig:rsquares}, we plot the histograms for $R^2$ values computed when the meshes are formed using (a) random voxels, (b) spatially nearest voxels,  and (c) functionally nearest voxels.

In the experiments, if seed voxels of meshes formed using spatial or functional neighbors are represented in terms of their neighbors, then we observe that the mean values of histograms for $R^2$ are $0.57$ and $0.54$, respectively (Fig. \ref{fig:rsquares}.b and c). On the other hand, if we represent seed voxels in terms of random voxels, then we observe larger estimation errors, and the mean value reduces to $0.29$ (Fig. \ref{fig:rsquares}.a). Therefore, employing spatial or functional neighbors during the construction of meshes results in better model fit compared to selecting random voxels.

\section{Conclusion}

In this study, we propose a method which maps a sequence of brain volumes, recorded during a cognitive stimulus to a  brain graph consisting of a set of local meshes. The proposed model, called ensemble of local meshes, defines two types of local meshes around each voxel, namely spatially local mesh (SLM) and functionally local mesh (FLM). While SLM models the relationship among voxel BOLD responses in a spatial neighborhood system, FLM models the relationship among voxel BOLD responses in functional neighborhood defined by Pearson correlation. We have observed that are considering the whole BOLD responses recording during a cognitive stimulus for the employment of spatial and functional relationship among voxels enables us to model the discriminative information in fMRI data. When we represent the brain encoding and decoding precess by a machine learning algorithm, namely Support Vector Machines (SVM), we observe that  \textbf{{FLM}} features perform substantially better among the features that use the state-of-the-art MVPA models.

We have observed that classification performances of SVM depend on the mesh size $p$, and  varies for each participant. We have also observed that employment of temporal measurements for modeling of mesh edge weights results in less standard deviation over various mesh sizes. Therefore, models which employ temporal measurements are more robust to the effect of using different mesh sizes for the extraction and classification of features. A future research direction will be the development of algorithms to generalize the ensemble of local meshes to model resting state or disease fMRI data. 

\section*{Acknowledgment}

We thank Orhan Firat, Baris Nasir, Arman Afrasiyabi, Burak Velioglu, Hazal Mogultay, Emre Aksan and Sarper Alkan for insightful discussions and fMRI data gathering. We also thank UMRAM for fMRI recordings. This work is supported by TUBITAK under the grant number 112E315.

% trigger a \newpage just before the given reference
% number - used to balance the columns on the last page
% adjust value as needed - may need to be readjusted if
% the document is modified later
%\IEEEtriggeratref{8}
% The "triggered" command can be changed if desired:
%\IEEEtriggercmd{\enlargethispage{-5in}}

% references section

% can use a bibliography generated by BibTeX as a .bbl file
% BibTeX documentation can be easily obtained at:
% http://www.ctan.org/tex-archive/biblio/bibtex/contrib/doc/
% The IEEEtran BibTeX style support page is at:
% http://www.michaelshell.org/tex/ieeetran/bibtex/
%\bibliographystyle{IEEEtran}
% argument is your BibTeX string definitions and bibliography database(s)
%\bibliography{IEEEabrv,../bib/paper}
%
% <OR> manually copy in the resultant .bbl file
% set second argument of \begin to the number of references
% (used to reserve space for the reference number labels box)
\bibliographystyle{IEEEtran}
% argument is your BibTeX string definitions and bibliography database(s)
\bibliography{jstsp_submitted}

% that's all folks
\end{document}